\documentclass{IEEEtran}
\IEEEoverridecommandlockouts
\usepackage{cite}
\usepackage{graphicx}
\usepackage{titlesec}
\titlespacing{\subsection}{2pt}{\parskip}{-\parskip}
\usepackage{subcaption}
\usepackage{cuted}
\usepackage{amsmath,amssymb,amsfonts}
\usepackage{textcomp}
\usepackage{xcolor}
\usepackage{amssymb}
\usepackage{algorithm}
\usepackage{algorithmicx}
\usepackage[noend]{algpseudocode}
\usepackage{caption}

\begin{document}

\title{Adaptive ResNet Architecture for Distributed Inference in Resource-Constrained IoT Systems}
\author{\IEEEauthorblockN{Fazeela Mazhar Khan, Emna Baccour, Aiman Erbad and Mounir Hamdi}\\
\IEEEauthorblockA{College of Science and Engineering, Hamad Bin Khalifa University, Qatar Foundation, Doha, Qatar.}
}
\maketitle

\begin{abstract}
As deep neural networks continue to expand and become more complex, most edge devices are unable to handle their extensive processing requirements. Therefore, the concept of distributed inference is essential to distribute the neural network among a cluster of nodes. However, distribution may lead to additional energy consumption and dependency among devices that suffer from unstable transmission rates. Unstable transmission rates harm real-time performance of IoT devices causing low latency, high energy usage, and potential failures. Hence, for dynamic systems, it is necessary to have a resilient DNN with an adaptive architecture that can downsize as per the available resources. This paper presents an empirical study that identifies the connections in ResNet that can be dropped without significantly impacting the model's performance to enable distribution in case of resource shortage. Based on the results, a multi-objective optimization problem is formulated to minimize latency and maximize accuracy as per available resources. Our experiments demonstrate that an adaptive ResNet architecture can reduce shared data, energy consumption, and latency throughout the distribution while maintaining high accuracy.
\end{abstract}


\section{INTRODUCTION}
Artificial Intelligence (AI) plays a crucial role in handling the vast amounts of data that is being generated from Internet of Things (IoT) devices. By leveraging AI algorithms, IoT devices can provide better insights and have more intelligent decision-making capabilities. More specifically, Deep Neural Networks (DNNs) are known to be highly effective in solving complex problems in various fields, such as computer vision, natural language processing, etc. \cite{r1,r2}. However, they are resource-intensive, needing considerable computing power and energy for both training and operation. To deal with resource constraints in IoT devices, hosting DNNs in the cloud and sending IoT data for processing is an alternative. However, cloud deployment results in high network resource consumption increased latency, and potential privacy concerns due to continuous data transmission during inference.\cite{r3}.

To improve IoT with deep networks, researchers are creating methods for optimizing deep networks such as distributed inference, pruning, and quantization \cite{r4,r5}, and developing more efficient network architectures \cite{r7}. Although distributed inference can help IoT devices overcome some of the challenges associated with centralized processing, it does have potential issues to consider. Distributing inference tasks across multiple devices requires communication and synchronization between the devices, which can consume significant amounts of energy. Additionally, distributed inference introduces dependency between devices and has potential for unstable transmission rates due to mobility. These unstable transmission rates lead to low latency, higher energy consumption and potential failure,  which is not ideal for real-time IoT applications.

Data compression is a common technique used to reduce the amount of computed and transmitted data, saving bandwidth and computational resources. This can be achieved through different compression algorithms such as pruning, quantization, etc. \cite{r4,r5}. However, in IoT applications, the data generated can be highly dynamic, with significant variations in the load over time. This leads to one of the main issues with using compression, which is the non-dynamic nature of compression algorithms used in the literature. These algorithms require retraining before online use, which makes them not prone to online changes in order to be able to adapt to the dynamics of available IoT resources and rate of collected data. Furthermore, compression can have a negative impact on the accuracy of non-robust DNNs, particularly if the compression is aggressive or not tailored to the network and dataset. This impact cannot be improved even when sufficient resources are available due to the non-dynamic nature of most compression algorithms. 

Residual Network (ResNet) is a popular DNN architecture that improves accuracy and performance in various applications. It uses residual connections to pass information directly between layers, allowing the network to retain information about critical features. \cite{r6}. Skip hyperconnections are added to ResNet to integrate resiliency into the training process and achieve robustness in case of node failure or communication problems during distributed inference.\cite{r7}.

Considering the problems associated with compressing or distributing the DNN, we propose to design an adaptive ResNet architecture for distributed inference that ensures resiliency in case some computational tasks are skipped due to resource shortage. Additionally, we will customize the ResNet connections online in order to adapt to the dynamics of the devices and available resources during the distribution while ensuring to maintain the highest possible accuracy. It is important to consider skipping some connections in cases such as low transmission rates or battery draining because IoT devices are mostly resource constrained. Thus, while not impacting the latency of the inference, especially for time-sensitive applications, the proposed system ensures the highest possible accuracy as per the resource availability. This paper targets to do an empirical study in order to identify connections in the residual network that do not have a major impact on the accuracy as their effect is mitigated by the skip connections. On the basis of these results, the system is able to adapt the ResNet architecture as per the dynamics of devices, available resources, and the rate of collected data by skipping some connections and computational tasks. Moreover, the latency can be optimized if the system operator can sacrifice the accuracy.  The main contributions of this paper are:
\begin{itemize}
    \item Conducting an empirical study to identify the potential connections that can be dropped while the network’s performance is not impacted significantly. 
    \item Formulating a multi-objective optimization problem that targets minimizing the latency and maximizing the accuracy. The optimization is NP-hard and considers the dynamic properties of IoT devices.
    \item Relaxing the optimization to reach a sub-optimal heuristic that achieves the targets of our system.
    \item Evaluating our approach under different configurations of the network and parameters of the system.  
\end{itemize}

\begin{figure}[h]
  \centering
  \includegraphics[width=0.3\textwidth]{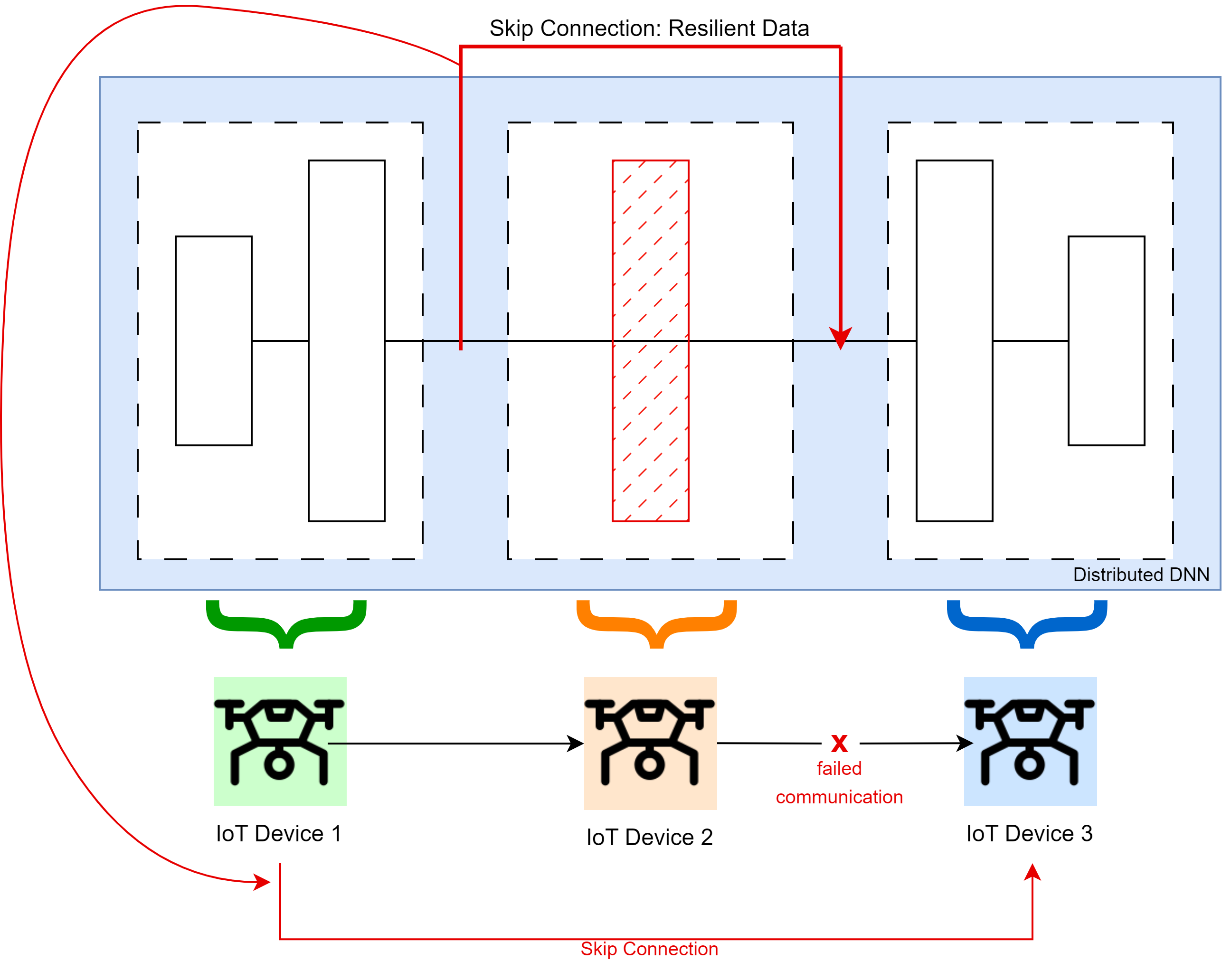}
  \caption{Illustration of System Model}
  \label{fig1}
\end{figure}
\vspace{-0.2cm}
\section{RELATED WORK}
As AI becomes more common in IoT devices, recent research focuses on executing inference tasks locally to reduce latency. Some use compression algorithms, while others use distributed inference. However, many pruning techniques require dedicated resources or significant fine-tuning \cite{r13}, making them unsuitable for IoT devices in online scenarios. This is because the pruned network size cannot adapt dynamically to available resources, and retraining the entire network is resource-intensive. As a result, only one network with reduced accuracy is usually deployed online.

In \cite{r8}, the authors propose a distributed CNN inference method such that although the devices involved have limited resources, they could still perform the inference. An optimization problem is formulated where the target is to minimize the latency. The same model also caters to the mobility aspect of the devices involved as it does have an impact on the latency. To emphasize the privacy aspect of distributed inference in IoT devices, \cite{r9} highlighted the concerns raised by sharing the data among untrusted devices. Specifically, an untrusted IoT ad hoc network device used for processing inference requests can reconstruct the complete original data from limited information. A joint inference model is proposed known as DistPrivacy, which respects the privacy aspect of the original data by distributing the feature maps into more devices to strengthen the privacy of the original input.  A survey \cite{r10} discusses that when performing distributed inference, it is crucial to consider the size of AI models since their computational and energy requirements may surpass the capacity of some devices. Additionally, decentralized training or inference is challenging due to wireless communication between participants, causing issues such as network capacity, performance delays, and privacy concerns. Consequently, AI implementation requires careful consideration of various factors such as partitioning the model, efficient architecture and algorithms, and communication protocols while also respecting the on-device constraints.

\cite{r7} explores the resilience of DNNs in the event of physical node failure or failure of DNN units on those nodes. The method, namely \textit{deepFogGuard} successfully implements the concept of skip hyperconnections, to ensure inference resiliency is developed and maintained in case of node failures. \textit{deepFogGuard} introduces awareness amongst the nodes such that in case of node failure, the next physical node can to detect the failure. This can be done through a keep-alive mechanism to maintain an active connection between devices by sending periodic messages to prevent timeout or termination. Although the proposed method improved the model’s resilience against physical node failures it is still limited by the heavy amount of bandwidth consumption needed for skip hyperconnections. To limit the bandwidth consumption, \cite{r6} investigates skip hyper-connections to identify their relative importance.

As the previous work focuses more on layer-by-layer implementation thus, the proposed system model focuses on a block-by-block implementation because residual blocks are an important aspect of ResNets.  A residual block in ResNet consists of multiple convolutional layers with batch normalization and ReLU activation functions, and a skip connection that adds the output of previous block to the output of the current block before applying the final activation function. The skip connections in each residual block of ResNet help alleviate the vanishing gradient problem and enable direct information flow, leading to the learning of more fine-grained features.  Additionally, since IoT devices are always moving and the collected data rate is volatile,  we consider it is essential to consider the dynamic properties of such mobile environments such as unstable transmission rates which have an impact on the latency and energy of the system. Hence, we propose to design an adaptive ResNet with distributed architecture to adapt to the network dynamics.

\section{RESILIENCY OF RESNET ARCHITECTURE}
ResNet (short for Residual Neural Network) is a deep learning architecture that was first introduced in 2015 by He et al. as a solution to the problem of vanishing gradients in very deep neural networks. ResNets have been widely adopted and have achieved state-of-the-art results on many image classification benchmarks. One of the main features of ResNets is the use of residual connections, which enable the network to be very deep while maintaining good performance.

  In this section, we present our system where the architecture of ResNet is adaptive to the dynamic aspect of Resource Constrained IoT devices in order to assist the distributed inference while having a resilient accuracy. We start with identifying the impact of different ResNet blocks on the performance of the network. The results of this empirical study are then used in the optimization so that the system can adapt as per the available resources and if needed, can drop some blocks while slightly sacrificing the accuracy. Additionally, our proposed system also allows minimizing the latency, if the accuracy can be relaxed. 

\begin{figure}[htbp]
  \centering
  \includegraphics[width=0.4\textwidth]{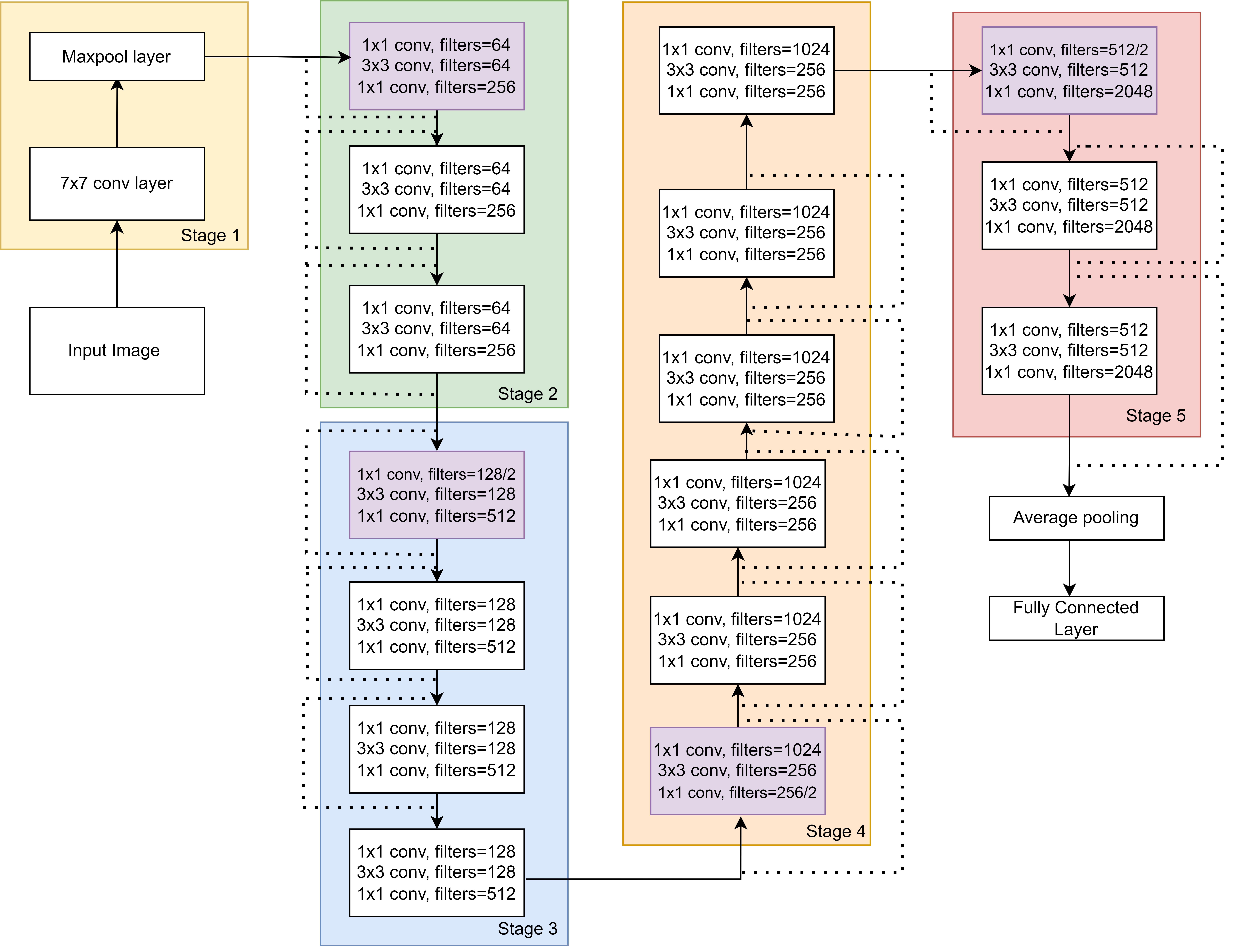}
  \caption{Block Implementation of ResNet-50.}
  \label{fig2}
\end{figure}

\begin{figure}[htbp]
  \centering
  \includegraphics[width=0.3\textwidth]{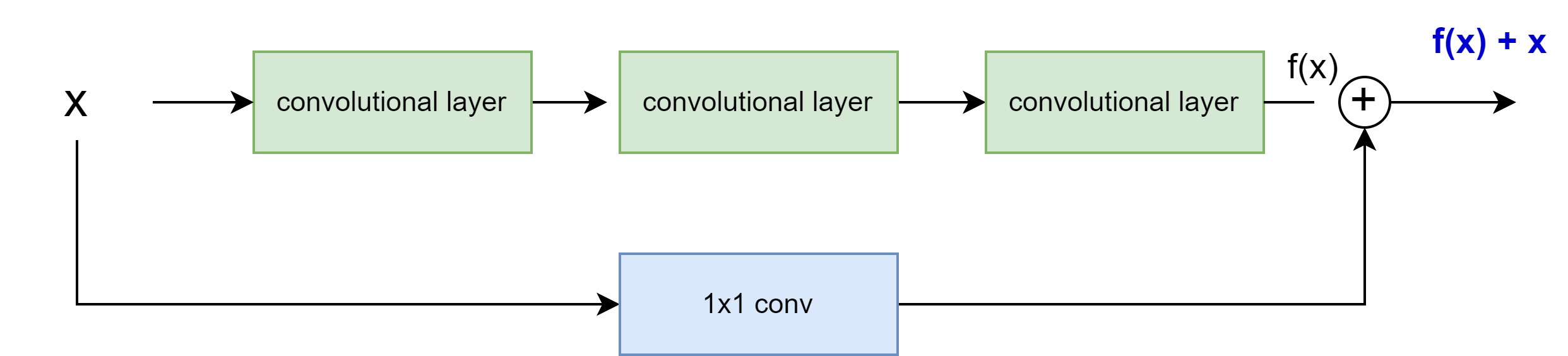}
  \caption{Convolutional Block}
  \label{fig3}
\end{figure}

\begin{figure}[htbp]
  \centering
  \includegraphics[width=0.3\textwidth]{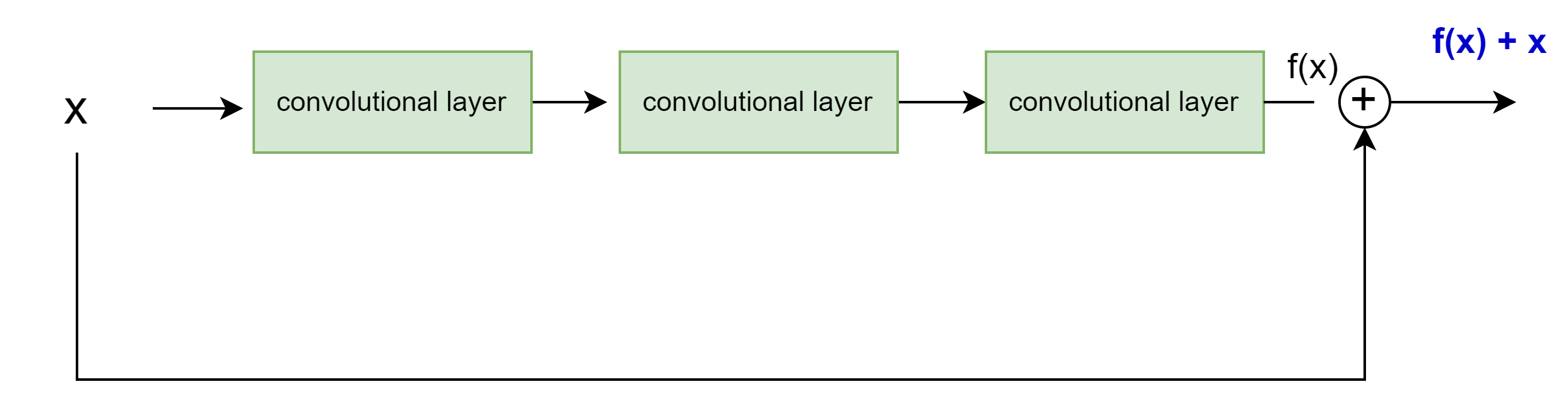}
  \caption{Identity Block.}
  \label{fig4}
\end{figure}

\begin{figure}[htbp]
  \centering
  \includegraphics[width=0.25\textwidth]{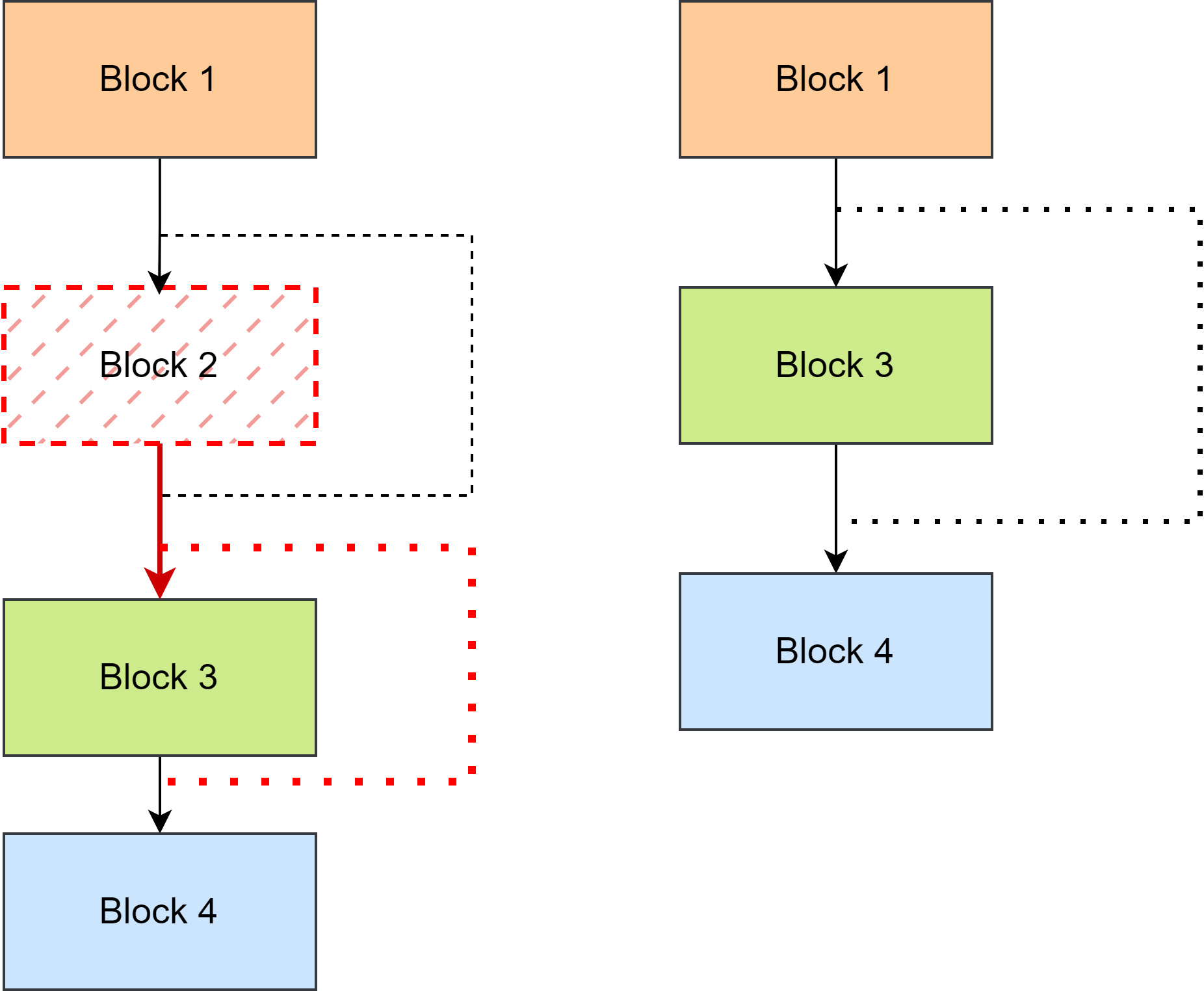}
  \caption{Illustration of a dropped block}
  \label{fig5}
\end{figure}

Without loss of generality, we conducted our study on ResNet-50 as it has been widely used for complex image classification tasks, achieving state-of-the-art performance on several benchmark datasets such as CIFAR-10 \cite{r14}. A block-by-block implementation as shown in Figure \ref{fig2} of the ResNet-50 model is executed to train the model on CIFAR-10 dataset. The dataset consists of nearly 50,000 training images and 10,000 test images split into 10 categories.  We fine-tune ResNet-50 with a batch size equal to 512, a Stochastic Gradient Descent (SGD) optimizer with a momentum of 0.9, a learning rate equal to 0.01, to achieve a test accuracy of 94.73\%. The model being used is a pre-defined ImageNet model which is being trained with SGD as the loss function for 30 epochs. The input image size is upsampled to 224x224x3. In order to study the effect of dropping certain connections, the model is implemented using stages and blocks. In ResNet-50, a stage is a group of convolutional layers that perform a specific set of operations. The first stage of ResNet-50 consists of a 7x7 convolutional layer with 64 filters, followed by a batch normalization layer, a ReLU activation function, and a 3x3 max pooling layer.The remaining stages consist of a series of bottleneck blocks, which use 1x1, 3x3, and 1x1 convolutions. The first block of each stage is known as  convolutional block, as the input size is not equal to the output size of the block and can be defined in Figure \ref{fig3} where x is the input to the block. The blocks that follow in figure \ref{fig4} have an input size equal to the output size and are known as identity blocks. The main target when evaluating the resiliency of the network is to drop certain blocks in order to identify the impact of dropped connections on the network’s accuracy. When a block is dropped, the implementation is as if the block was never computed. Additionally, the first stage is the input stage and thus, is essential to the network and cannot be dropped. The inputs to the following block can be explained as in figure \ref{fig5}. 

The results of the empirical study are illustrated in Table \ref{tab1} where first, both types of connections: Direct Connections (DC) and Skip Connections (SC) are dropped. The results prove that SC are more important for the network's performance which is justified as SC help to preserve information and prevent loss of accuracy. Thus, we choose to do a deeper study on the dropping of DC trying at most of two blocks being dropped at a time because more dropped blocks result in unacceptable accuracy as more than 6 convolutional layers are skipped. As a result, we conclude the results of the empirical study in table \ref{tab2}, where we consider only the scenarios where the accuracy is more than 80\% while calculating the memory and computation gain by skipping the block(s). These results are to be utilized in the next section. 
    
 \begin{table}[htbp]
  \centering
  \renewcommand{\arraystretch}{0} 
  \begin{tabular}{@{}c@{}}
    \includegraphics[width=0.4\textwidth]{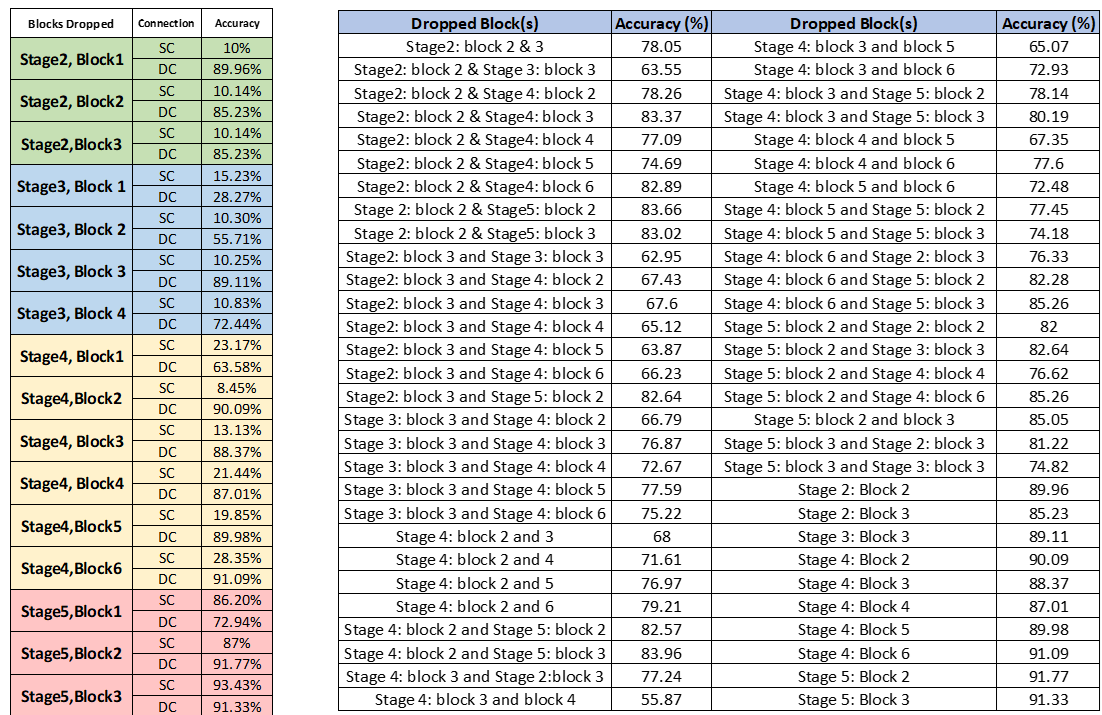} \\
  \end{tabular}
  \caption{Results of the Empirical Study}
  \label{tab1}
\end{table}

    \begin{table}[htbp]
  \centering
  \renewcommand{\arraystretch}{0} 
  \begin{tabular}{@{}c@{}}
    \includegraphics[width=0.3\textwidth]{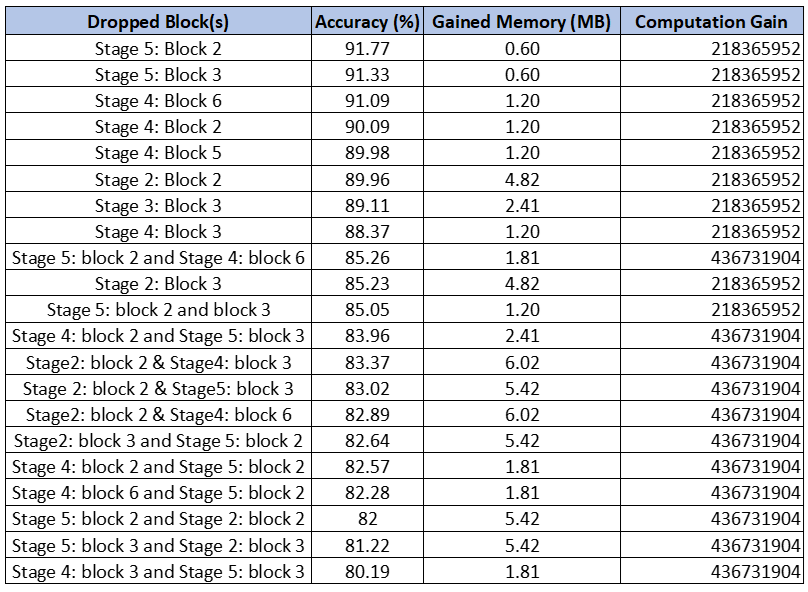} \\
  \end{tabular}
  \caption{Concluded Results of the Empirical Study}
  \label{tab2}
\end{table}

\section{Adaptive ResNet Architecture for Distributed Inference}
In this section, we present our system model that is adaptive to the changes in the architecture of ResNet, in order to adapt to the dynamic aspect of Resource Constrained IoT devices.
\subsection{System Model}
 The proposed system model comprises of implementing a distributed ResNet network, where blocks are allocated across multiple IoT devices with skip connections for resilient inference as shown in figure \ref{fig1} We formulate our approach as an optimization problem that targets to minimize the latency and maximize the accuracy, while also respecting the available IoT resources and the required accuracy threshold. The system model consists of \textit{N} devices, each has a memory limit of $\overline{m_i}$, computational limit of $\overline{c_i}$, energy capacity of  $\overline{e_i}$ and \textit{e(i)} is device \textit{i}'s limit for the number of multiplications it can perform in a second. \textit{M} is the total number of blocks per network, $\rho_{(i,j)}$ is the data rate between two devices $i$ and $j$, for a total of $R$ requests. The mobility of these devices is indirectly studied through the variation of $\rho$.

Equation \ref{(5)} describes the memory requirement of a total of \textit{L} layers in block \textit{j} $ \in \{1, . . . , M\}$,  where $W_{j,l}$ is the size of the inputs of layer in the block \textit{j} for each layer \textit{l}, where $\l$ $ \in \{1, . . . , L\}$ , and \textit{b} is equal to 4 bytes required to store each weight of the data. The memory load of all layers is  then added to finally compute the memory load of each block. Similarly, the computational requirements of each block  $c_j$ can be calculated through Equation \ref{(6)} where $n_{(k-1)}$ is the filter size for the previous layer \textit{k},  $s_k$  is the kernel size,  $n_k$  is the filters size of layer \textit{k}, and $O_k$  is the input shape as defined in \cite{r9}. 

\begin{equation}\label{(5)}
m_j= \sum_{l=1}^L W_l.\ b  
\end{equation}
\begin{equation}\label{(6)}
c_j= \sum_{l=1}^L n_{(k-1)}.\ s_k\ \ .\ n_k\ \ .\ O_k 
\end{equation}

\subsection{Problem Formulation}
The optimization problem is multi-objective and targets minimizing latency while maximizing accuracy. It depends on two decision variables which are the allocation decision variable defined in Equation \ref{(7)} and the model downsizing decision variable defined in Equation \ref{(8)}.

\begin{align}\label{(7)}
    x_{r,i,j} = 
    \begin{cases}
        1 & \text{if device \textit{ i} executes block \textit{j} of request \textit{r},} \\
        0 & \text{otherwise.}
    \end{cases}
\end{align}

\begin{align}\label{(8)}
    y_{r,j} = 
    \begin{cases}
        0 & \text{drop block \textit{j} of request \textit{r},} \\
        1 & \text{otherwise.}
    \end{cases}
\end{align}

The objective function defined in Equation \ref{(12)} consists of two weighted variables $\alpha$ and $\beta$ which allow prioritizing either latency or accuracy.

\begin{equation}\label{(12)}
\centering
  W_o=\ \alpha\ .\ \ \frac{\overline{T_L}}{R}+\ \beta\ .\ ({1-\ \ Acc})
\end{equation}

The first part of $W_o$ presents the average normalized latency $\overline{T_L}$. The second part represents the expected average accuracy loss of ResNet after potential dropping of some blocks, which is calculated using function $g(y_r)$ that provides the accuracy based on the decision variable $y_r$ and the empirical study conducted earlier. $y_r$ denotes the downsizing decisions of all the blocks of request \textit{r}.  Therefore, the accuracy is calculated as follows:
\begin{equation}\label{(11)}
\centering
Acc= \sum_{r=1}^R \ \cdot \  \frac{g(y_{r})}{R}
\end{equation}

The objective function establishes a balance between the latency and accuracy (such that $\alpha$ + $\beta$ = 1). Accordingly, if $\alpha$ $>$ $\beta$ , minimizing the latency is a priority over the accuracy, as long as the tolerated accuracy is achieved. If  $\beta$ is equal to 0, maximum model downsizing is expected. If  $\alpha$ is equal to 0, the system resorts to downsizing only if resources are not  available.

\begin{equation}\label{(9)}
\centering
t_i = \sum_{r=1}^R \ x_{(r,i,j)}  . y_{r,j} \ . \ t_{(i,j)}^{(c)} 
\end{equation}

\begin{equation}\label{(10)}
\centering
\begin{aligned}
\hspace{0.2cm}T_L &= \sum_{r=1}^R \sum_{i=1}^N\ \ \sum_{j=1}^M  \sum_{k=1\ ,\ k\neq i}^N \ \sum_{n=1\ }^N \sum_{\sigma=1}^O   
\\
&\hspace{0.2cm}\max(x_{(r,i,j)} \ .\ x_{(r,k,j+1)} \ .\ t_{(i,j)}^{(tr)} \ .\ y_{(r,j)} , 
\\
&\hspace{0.2cm}x_{(r,k,j+1)} .\ x_{(r,n,j-\sigma+1)}.t_{(k,n,j,\sigma)}^{(sc)} \ .\ y_{(r,j)} ) + t_i \ \ 
\end{aligned}
\end{equation}

 The total latency, as shown in Equation \ref{(10)}, accounts for all requests and all blocks. It is composed of the following:
\begin{enumerate}
  \item The Computation Latency $t_{i,j}^{(c)}$ \  per device, is the time needed to compute block \textit{j} by device \textit{i} which is defined in  Equation \ref{(1)}. We utilize this for the total computation latency $t_i$ of the system, as defined in Equation \ref{(9)} which is the time for an IoT device to compute all assigned blocks based on the ratio of its computational load ($c_{j}$) to its multiplicative capacity per second ($e_i$). 
  \begin{equation}\label{(1)}
t_{(i,j)}^{(c)}= \frac{c_j}{e_i}
\end{equation}

  \item 	$t_{i,j,k}^{(tr)}$ is the transmission latency  between device \textit{i} and device \textit{k} for the direct connections communicating the output of block \textit{j}. $\rho_{i,k}$ is the data rate between the two devices and it is different for each device as the devices are constantly moving and do not have a fixed distance between them at all times. Lastly, $K_j$ is the output size of block \textit{j}.
  
\begin{equation}\label{(2)}
t_{(i,j,k)}^{(tr)}=\frac{K_j}{\rho_{(i,k)}}
\end{equation}
  \item Regarding the transmission latency of skip connections data $t_{k,n,j,\sigma}^{(sc)}$, it can be defined as in Equation \ref{(3)}, which is between device \textit{k} and \textit{n}, for block \textit{j} and \textit{j}- $\sigma$. Each residual block receives an input from the previous direct connection and another from the skip connection. Thus, $K_{j-\sigma}$ is the output size of block \textit{j} - $\sigma$ and $\rho_{k,n}$ is the data rate between the two devices. Lastly, $\theta$ is equal to 1 if block \textit{j}- $\sigma$ and block \textit{j} are connected through a skip connection and 0 otherwise, while $\sigma$ $ \in \{1, O\}$ where O is the maximum number of skipped blocks.
  \begin{equation}\label{(3)}
t_{(k,n,j,\sigma)}^{(sc)}=\frac{K_{(j-\sigma)}}{\rho_{(k,n)}} \ . \ \theta_{(j,j-\sigma)}
\end{equation}

\end{enumerate}

The optimization defined in \ref{(00}, targets to minimize the objective function to ensure minimum latency and maximum accuracy possible while respecting the constraints of the system model and the devices involved. 

\begin{equation}\label{(00}
\centering
\min(W_o)
\end{equation}

\textbf{s.t.}

\begin{equation}\label{(13)}
\centering
  \forall i \in \mathbb{N}_N \  \sum_{r=1}^R   \sum_{j=1}^M x_{(r,i,j)} \ . \ m_j \ . \ y_{(r,j)}  \leq \overline m_i
\end{equation}

\begin{equation}\label{(14)}
\centering
   \forall i \in \mathbb{N}_N \  \sum_{r=1}^R   \sum_{j=1}^M \ x_{(r,i,j)} \ . \ c_j \ . \ y_{(r,j)}  \leq \overline c_i
\end{equation}

\begin{equation}\label{(16)}
\centering
   \forall j \in \mathbb{N}_M , \ \forall r \in \mathbb{N}_R \ \ \sum_{i=1}^N \ x_{(r,i,j)} \ = 1
\end{equation}

\begin{equation}\label{(17)}
\centering
  \forall r \in \mathbb{N}_R \hspace{1cm}  g(y_{(r)}) \ \geq threshold 
\end{equation}
\begin{equation}\label{(15)}
\centering
\forall i \in \mathbb{N}_N \ T_{e^i}\leq \overline{e_i}
\end{equation}

\begin{equation}\label{(last)}
\centering
x,y \in \{0,1\}
\end{equation}

The equations that follow describe the constraints of the system where Equations \ref{(13)} and \ref{(14)} ensure that the device’s memory and computational are sufficient. Equation \ref{(15)} ensures the device's energy capacity is sufficient where $T_{e^i}$ is the total energy as defined in Equation \ref{(4)}, $P_c$ is energy consumption per second of computation and $P_t$ is energy consumption per second of data transmission between the two devices:

\begin{equation}\label{(4)}
 T_{e^i} = \sum_{j=1}^M \sum_{k=1}^N \ \sum_{\sigma=1}^O t_{(i,j)}^{(c)}\ . \ P_{c} \ + \ t_{i,j,k}^{(tr)} \ . \ P_{t} \ + \ t_{i,k,j,\sigma}^{(sc)} \ . \ P_{t} 
\end{equation}

Equation \ref{(16)} is used to ensure there is no redundancy in the system and that each block is computed by one device only. Equation \ref{(17)} ensures that the expected accuracy of each inference request does not drop below a certain threshold. The optimization process involves static parameters for various IoT devices, including the data rate $\rho$. Hence, to account for network variation and devices mobility, the optimization is executed at different time steps t=1:T. Lastly, Equation \ref{(last)} ensures the decision variables are binary.
\\
\subsection{Sub-Optimal Relaxation}
Even though Equation \ref{(16)} ensures the redundancy is avoided in the system, it is crucial to consider that this makes the optimization a NP-hard problem, which no known optimizer is capable enough to solve. Thus, we will relax this constraint as shows in Equation \ref{(18)} in order to solve it using a typical numerical tools such as MATLAB and achieve sub-optimal results. 

\begin{equation}\label{(18)}
\centering
 \forall j \in \mathbb{N}_M , \ \forall r \in \mathbb{N}_R \ \ \sum_{i=1}^N \ x_{(r,i,j)} \ \geq 1
\end{equation}

However, we still want to ensure that each request is assigned to only one device. This is done through algorithm \ref{algo_device_allocation}. The device allocation is dependent on choosing the device with a higher data rate in case multiple devices are assigned for inference. Lastly, for the execution of first layer if all devices have the same data rate limit, then they are assigned on the basis of higher capacity. 
\begin{algorithm}
  \tiny
  \caption{Device Allocation}
  \label{algo_device_allocation}
  \begin{algorithmic}[1]
  \For{$ t = 1$ to $T$}
    \For{$r = 1$ to $R$}
    \State $dvc \gets 0$
      \For{$k = 1$ to $M$}
        \State $no.OfAllocatedDvcs \gets 0$
        \State $devices \gets []$
        \For{$i = 1$ to $N$}
          \If{$x(r,i,k) = 1$}
            \State $devices \gets [devices, i]$
            \State $no.OfAllocatedDvcs \gets no.OfAllocatedDvcs + x(r,i,k)$
          \EndIf
        \EndFor
        \If{$no.OfAllocatedDvcs > 1$}
          \If{$any(devices == dvc)$}
            \For{$l = 1$ to $\text{length}(devices)$}
              \If{$devices(l) \neq dvc$}
                \State $x(r,devices(l),k) \gets 0$
              \EndIf
            \EndFor
          \Else
            \If{$dvc \neq 0$}
              \State $rhos \gets []$
              \For{$l = 1$ to $\text{length}(devices)$}
                \State $rhos \gets [rhos, \rho(dvc,devices(l))]$
              \EndFor
              \State $[xx,yy] \gets \text{max}(rhos)$
              \State $dvc \gets devices(yy)$
              \For{$l = 1$ to $\text{length}(devices)$}
                \If{$devices(l) \neq dvc$}
                  \State $x(r,devices(l),k) \gets 0$
                \EndIf
              \EndFor
            \Else
              \State $comp \gets []$
              \For{$l = 1$ to $\text{length}(devices)$}
                \State $comp \gets [comp, e(devices(l))]$
              \EndFor
              \State $[xx,yy] \gets \text{max}(comp)$
              \State $dvc \gets devices(yy)$
              \For{$l = 1$ to $\text{length}(devices)$}
                \If{$devices(l) \neq dvc$}
                  \State $x(r,devices(l),k) \gets 0$
                \EndIf
              \EndFor
            \EndIf
          \EndIf
        \EndIf
      \EndFor
    \EndFor
    \EndFor
  \end{algorithmic}
\end{algorithm}

\vspace{-0.2cm}
\section{EVALUATION}
In order to evaluate the proposed approach, this section investigates different scenarios and their impact on the performance of the system model. The implementation of the optimization is done on MATLAB using Genetic Algorithm Optimizer \footnote{https://www.mathworks.com/discovery/genetic-algorithm.html}. The impact of varying configurations of following variables on the total latency, accuracy, shared data between the devices to execute the inference, computational load and the amount of energy consumed is assessed. In order to make it a fair comparison and ensure the implementation caters to the dynamic rate of incoming requests, the Poisson distribution is used and the data rate of the transmission channel between devices $\rho$ follows a uniform distribution of [7.2,72.2] Mb/s.To simulate the variation of data rate experienced by different devices and given that we are constrained by the complexity of the optimization, we choose T=(Number of rounds) For the Total Energy, $P_c$ is 8W, and $P_t$ is 10W.
\\ 
\subsection{Investigating the impact of weighting variables: }
\begin{figure}[htbp]
  \centering
  \subfloat[Average Accuracy]{\includegraphics[width=0.45\columnwidth]{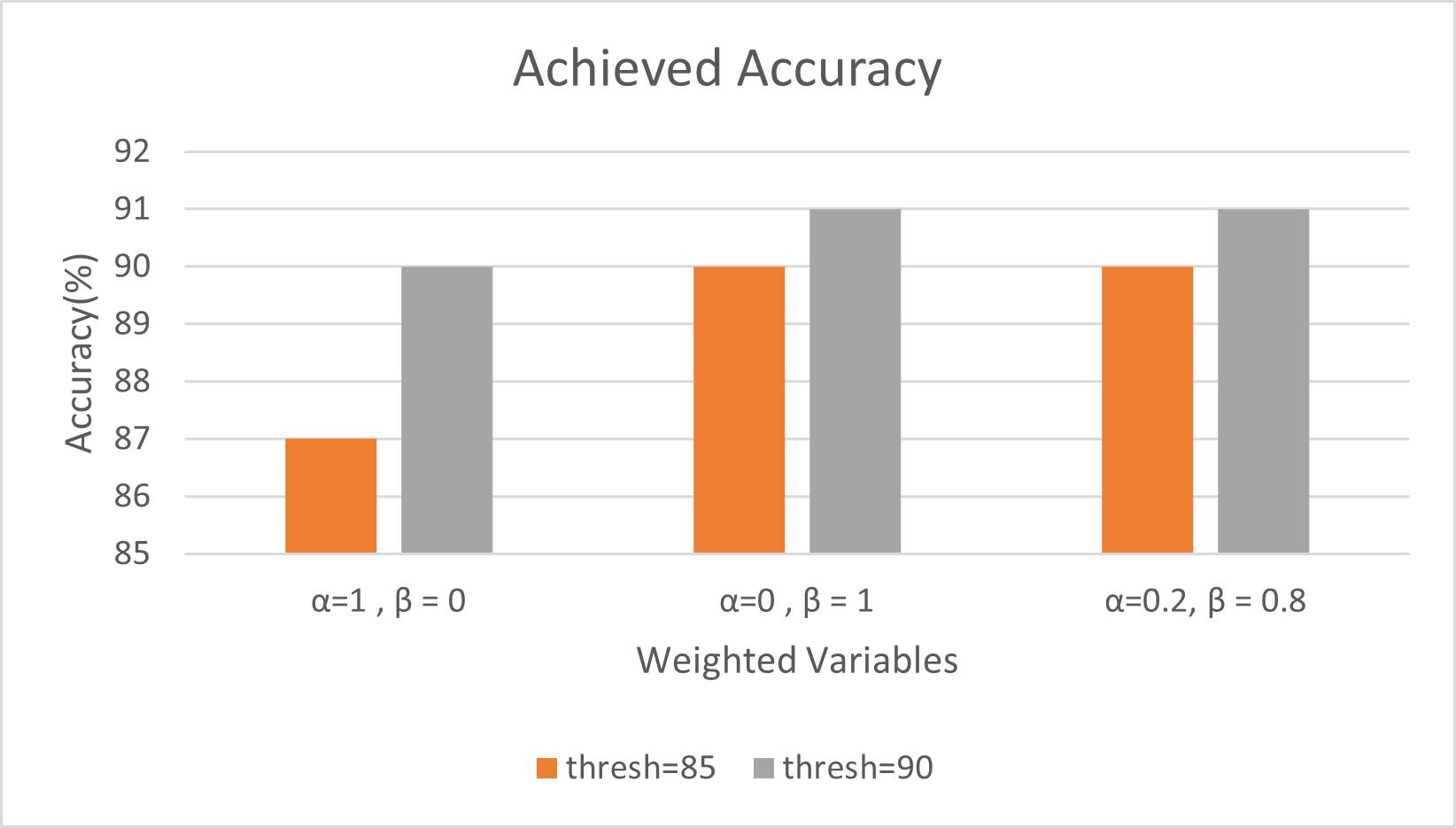}\label{fig:s1graph1}}
  \hfill
  \subfloat[Total Computation]{\includegraphics[width=0.5\columnwidth]{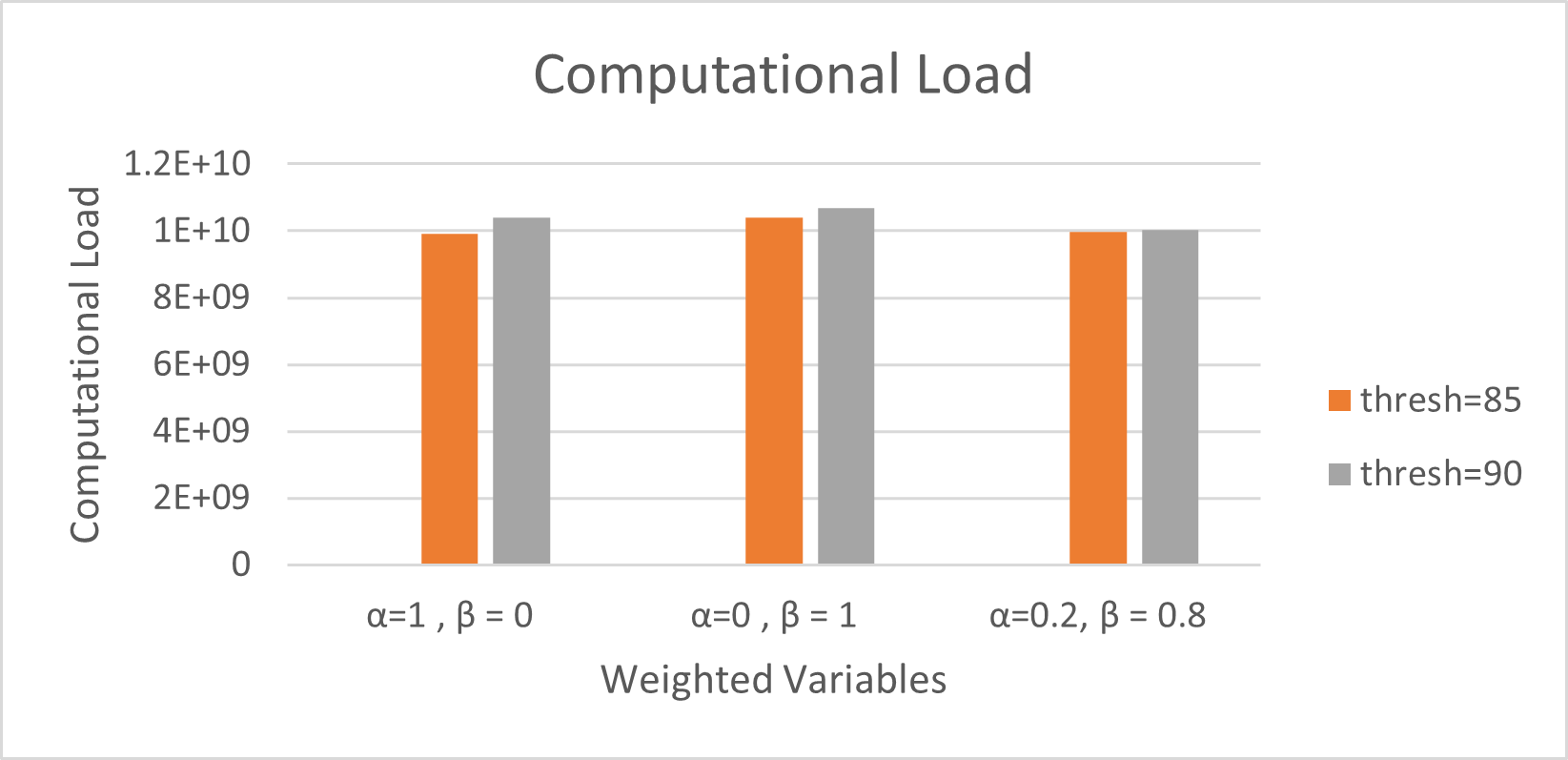}\label{fig:s1graph2}}
  \vfill
  \subfloat[Total Energy Consumption]{\includegraphics[width=0.45\columnwidth]{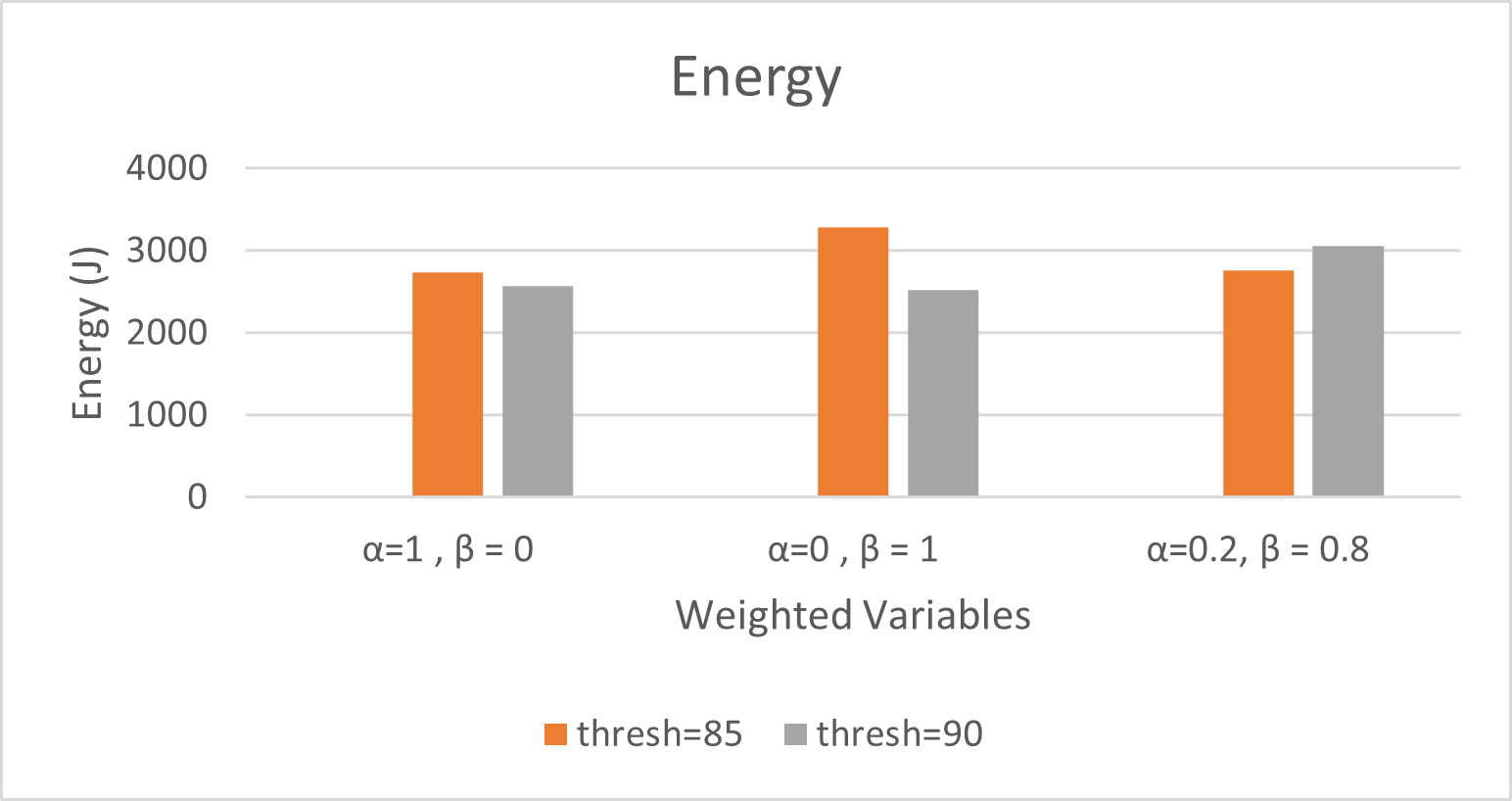}\label{fig:s1graph3}}
  \hfill
  \subfloat[Total Shared Data]{\includegraphics[width=0.45\columnwidth]{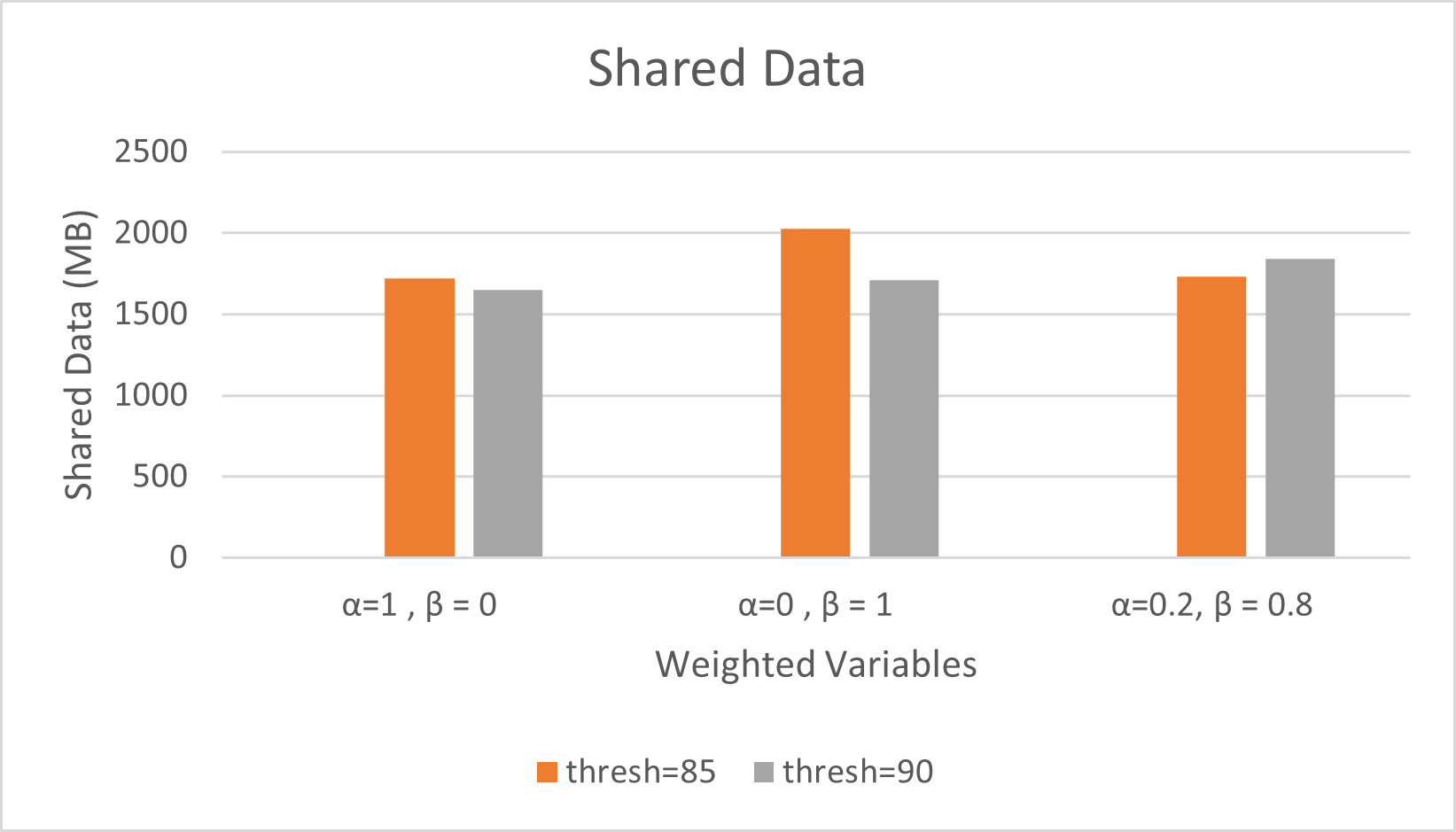}\label{fig:s1graph4}}
 
  \caption{Varying $\alpha$, $\beta$, and Accuracy Threshold of the system}
  \label{fig:Sc1}
\end{figure}

This section investigates the weighting variables $\alpha$  and $\beta$ for different accuracy thresholds. This allows us to explore the network's ability to maximize the accuracy and minimize the latency as per available resources and accuracy threshold. The number of IoT devices used is 10, which comprises devices with lower and higher capacities such that  $\overline{m_i} = $100MB and 200MB , $e(i)=$1.4 GF and 2.8 GF  , $\overline{e_i} = $1.4 GF, and 2.8 GF , $\lambda$ = 3 and  $\overline{c_i} = $800J and 1000J. The impact on accuracy is shown in figure \ref{fig:s1graph1} where the lowest accuracy recorded is when $\alpha = $ 1 and $\beta =$ 0 which proves that the system does not prioritize the accuracy beyond the defined threshold, and focuses on minimizing the latency.  Figure \ref{fig:s1graph2} shows that when $\alpha$ =1, less resources are needed but lower accuracy is achieved. Meanwhile, when $\beta$=1, the accuracy is the highest but higher resources are utilized to execute the inference. Additionally, the highest achieved accuracy when $\beta$=1 is 91\%, it means that the available resources are not enough to execute the inference without downsizing. Hence, despite the importance of the accuracy, the adaptive system executed the downsizing to accomplish the ResNet tasks.
\\
\begin{figure}[htbp]
  \centering
  \subfloat[Average Accuracy]{\includegraphics[width=0.45\columnwidth]{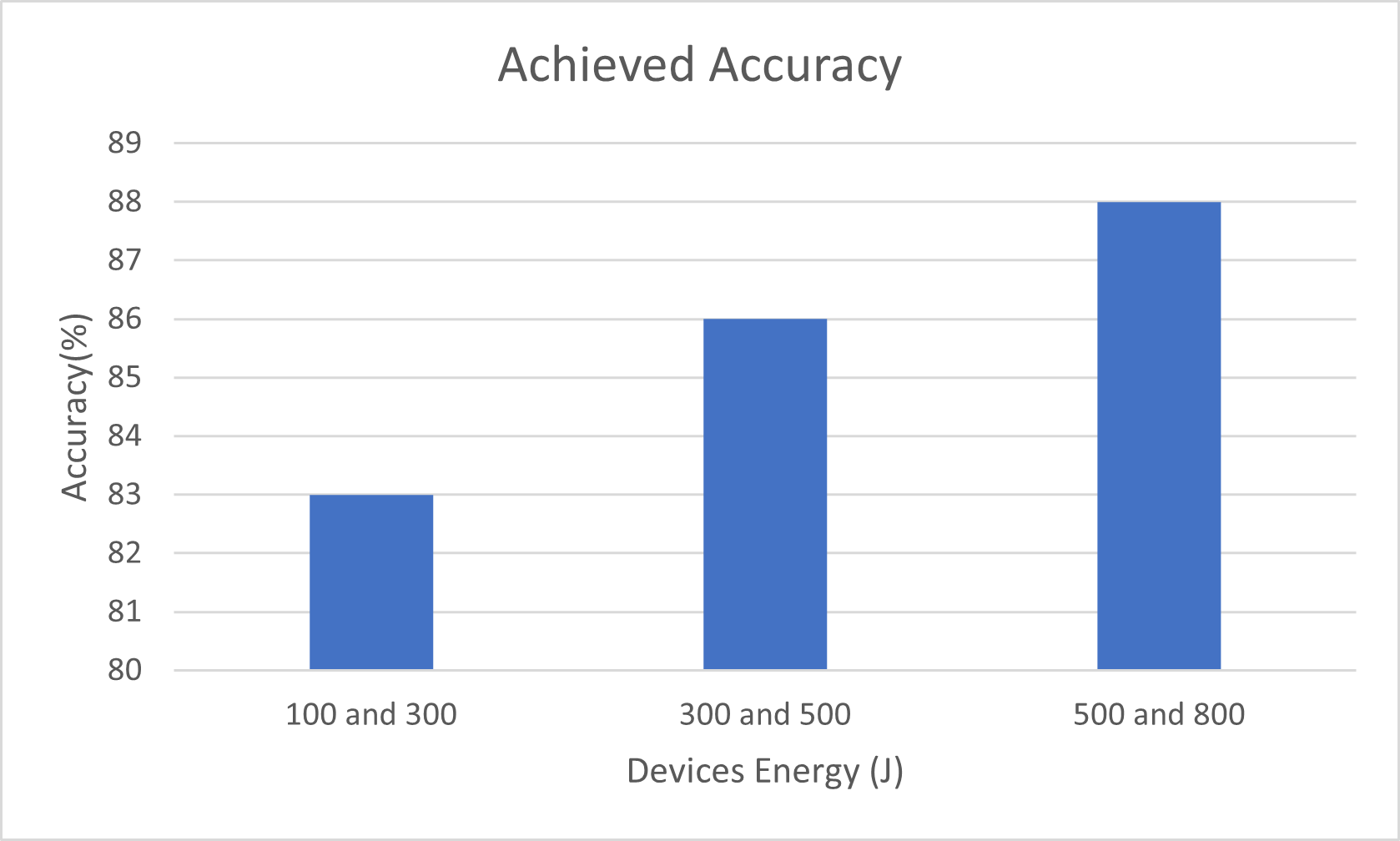}\label{fig:sc2graph1}}
  \hfill
  \subfloat[Total Latency]{\includegraphics[width=0.45\columnwidth]{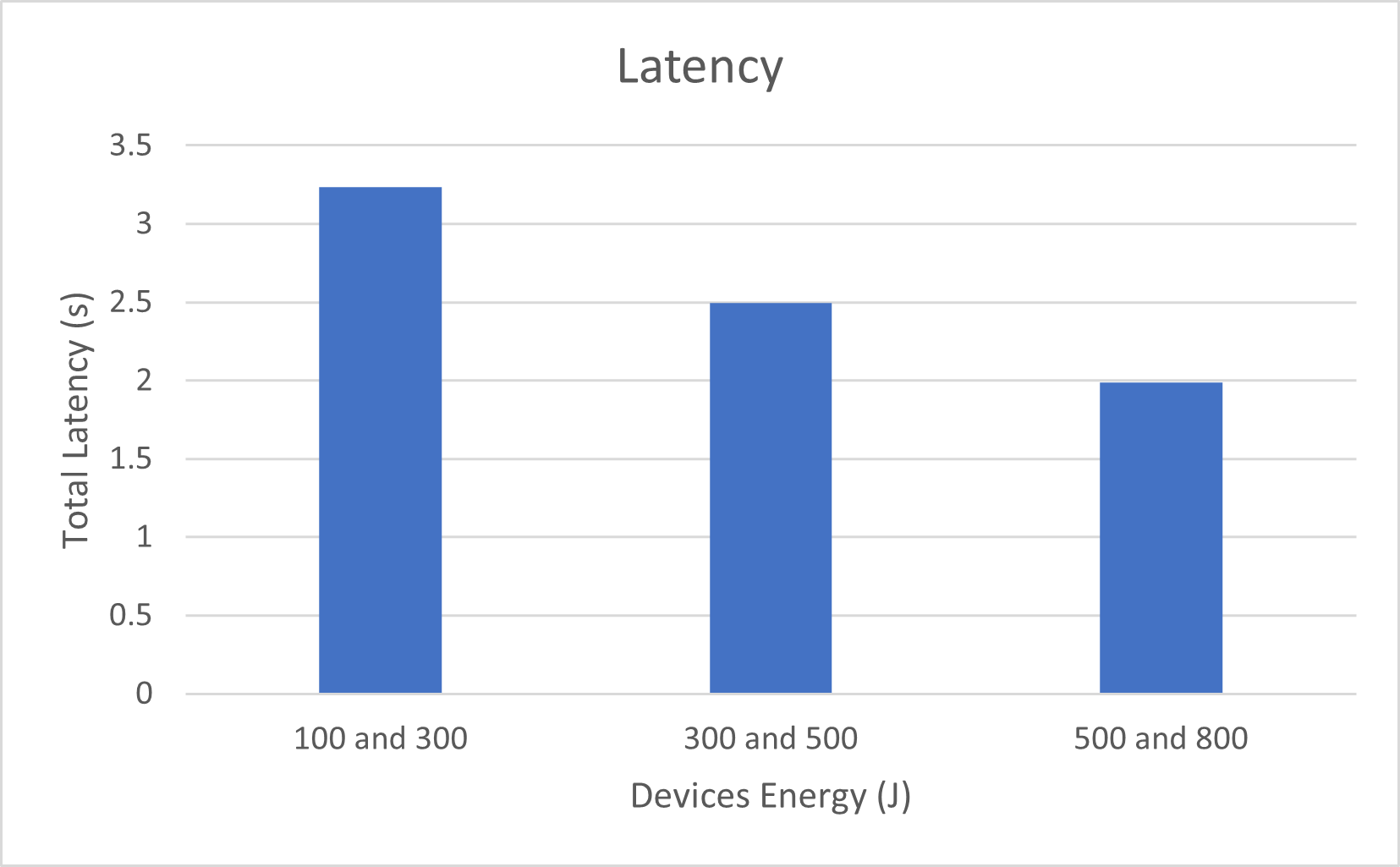}\label{fig:sc2graph2}}
  \vfill
  \subfloat[Total Shared Data]{\includegraphics[width=0.45\columnwidth]{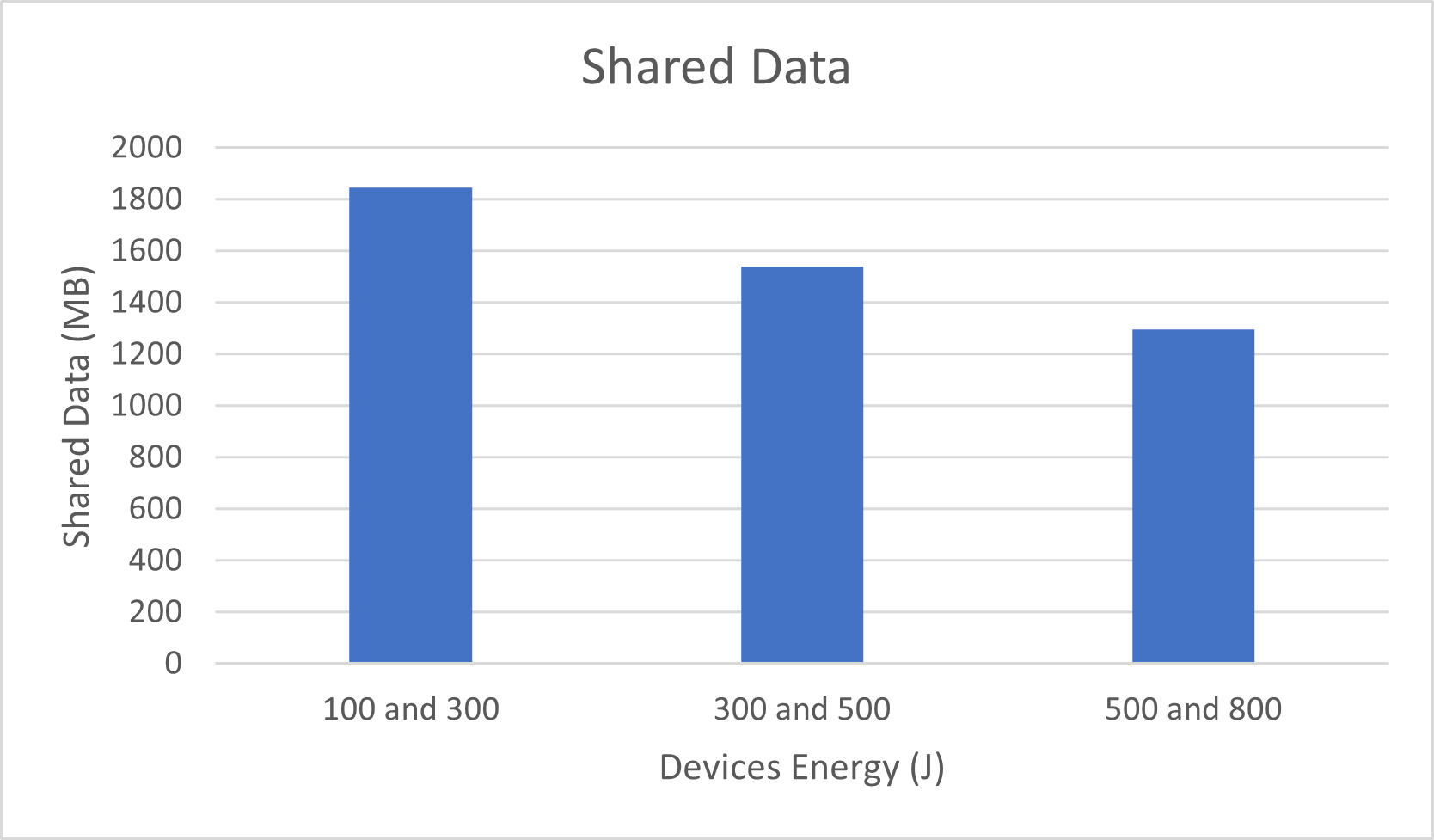}\label{fig:sc2graph3}}
  \hfill
  \subfloat[Total Computation]{\includegraphics[width=0.45\columnwidth]{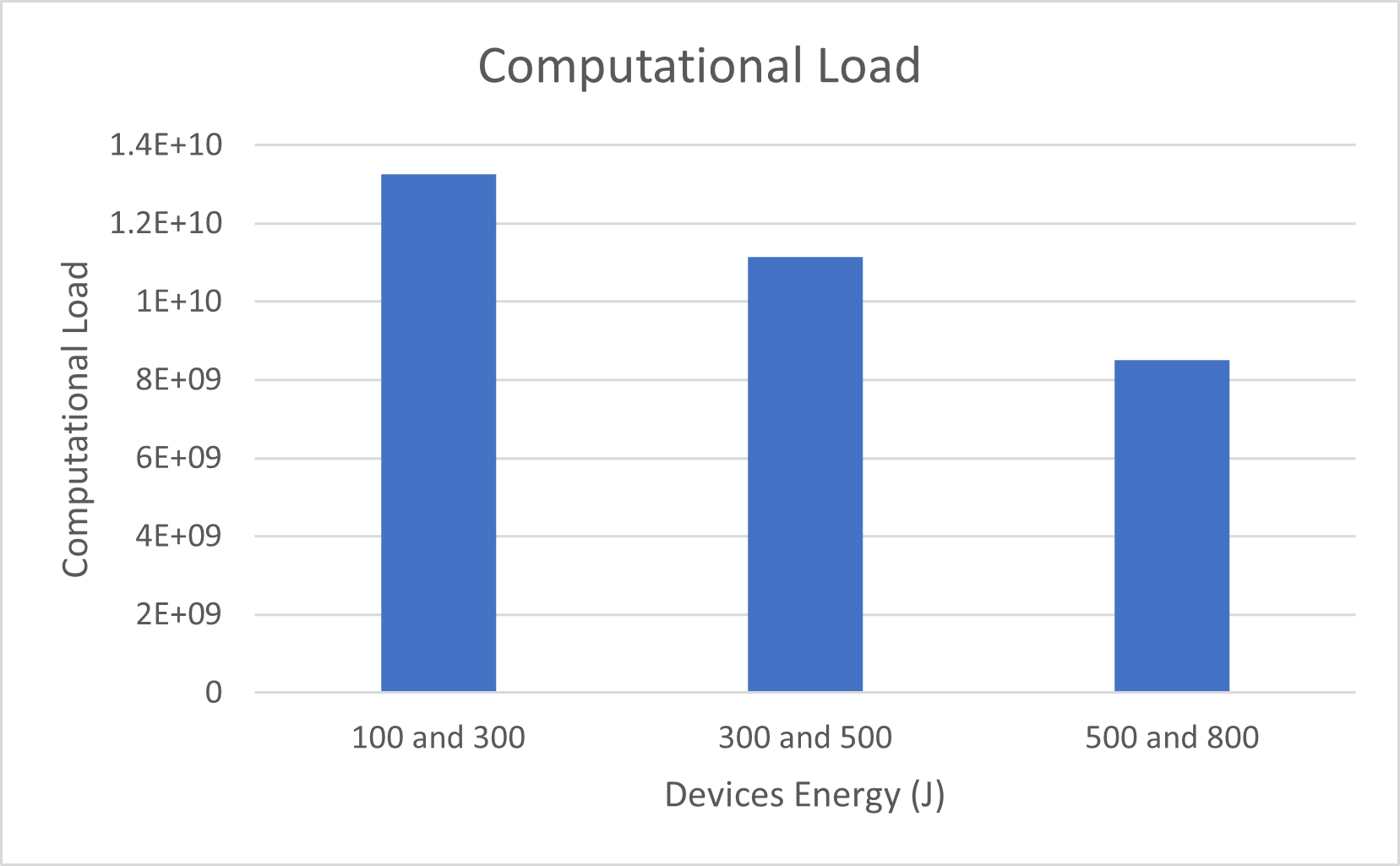}\label{fig:sc2graph5}}
  \vfill
    \subfloat[Total Energy Consumption]{\includegraphics[width=0.45\columnwidth]{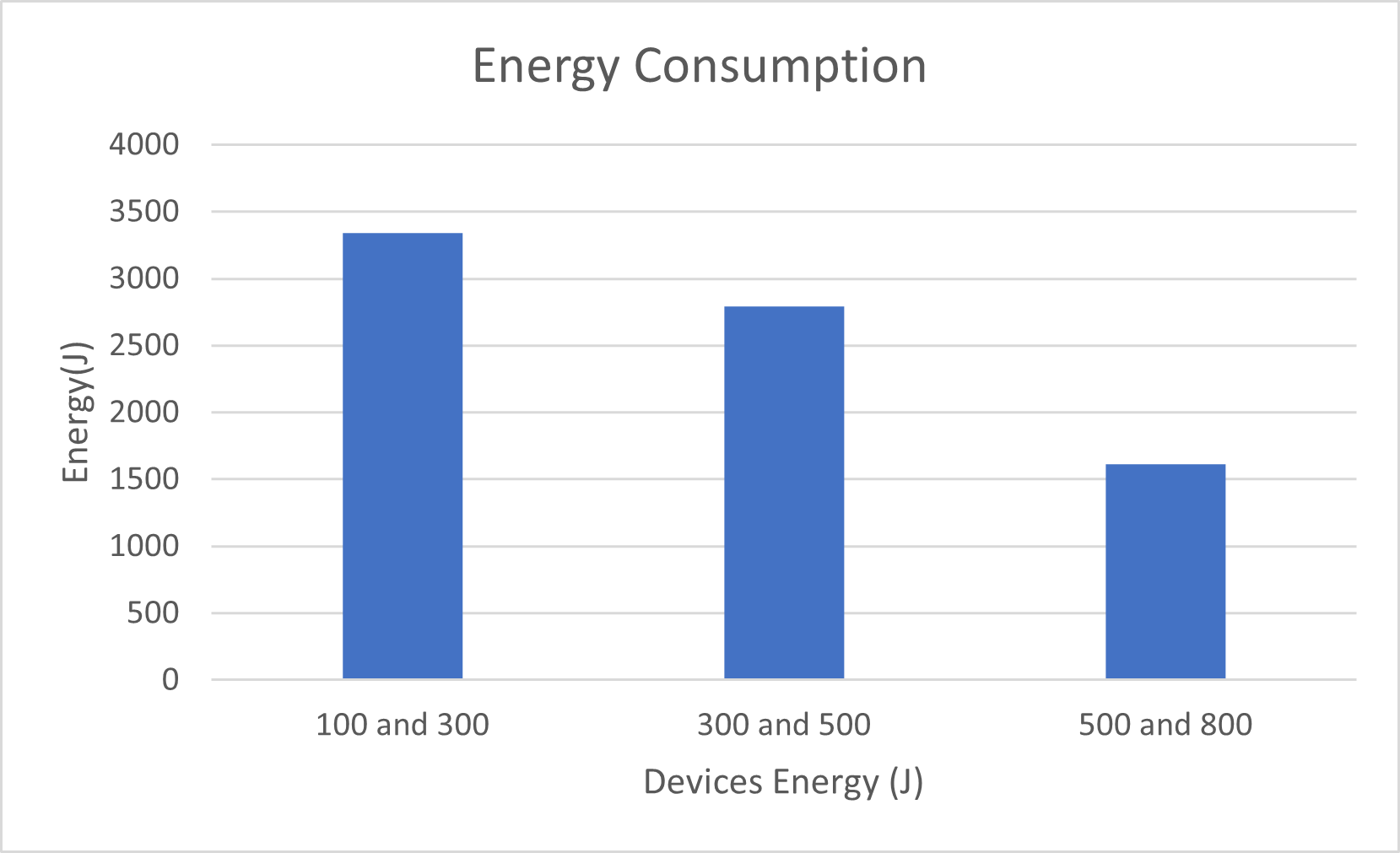}\label{fig:sc2graph4}}
  \caption{Varying the energy of the IoT devices}
  \label{fig:Sc2}
\end{figure}
\subsection{Investigating the change in battery capacity of devices: }
In order to assess the impact the batteries of devices have on the system, it is configured with same parameters as previous section with a threshold of 80\% while $\alpha$ = 0 and $\beta$ = 1. This is to investigate the proposed system model's capability to adapt the ResNet structure according to the available resources while maintaining the highest possible accuracy. figure \ref{fig:sc2graph1}  shows a clear trend in increased accuracy as the system's capacity increases and thus, it is able to perform better. This is because the system drops more blocks in case of a battery shortage, and thus the accuracy is reduced. On the other hand, when the battery is available within participants, the dropping is less aggressive. Since $\alpha$ = 0, the system does not prioritize on minimizing the latency. In summary, the latency decreases while the system's performance improves due to having sufficient resources as shown in \ref{fig:sc2graph2}. Specifically, more consecutive blocks can be computed in the same device and hence the transmission latency is reduced. Lastly, the shared data,  computational load and energy consumption for the resources decrease as shown in figures \ref{fig:sc2graph3}, \ref{fig:sc2graph5} and \ref{fig:sc2graph4}.
\\
\subsection{Investigating the change in the rate of Incoming Requests:}
\begin{figure}[htbp]
  \centering
  \subfloat[Average Accuracy]{\includegraphics[width=0.4\columnwidth]{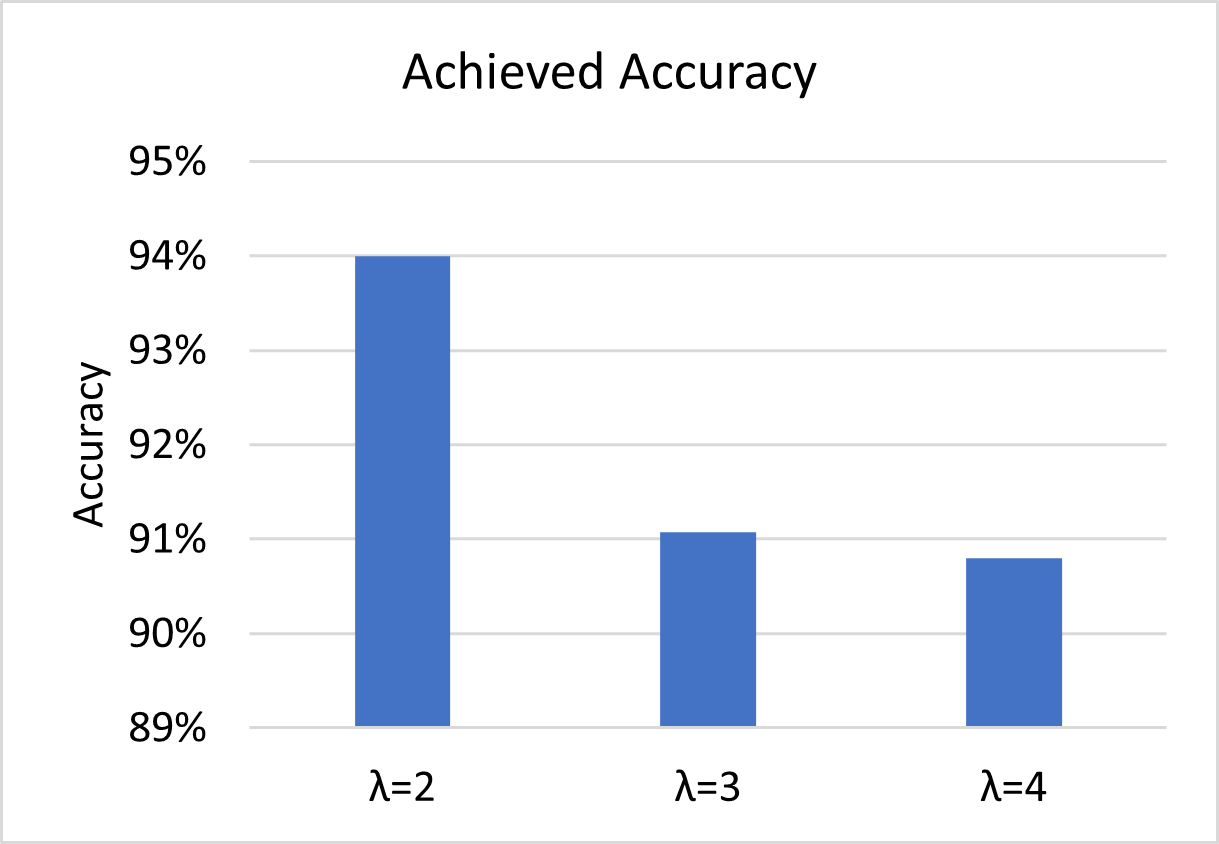}\label{figsc3:graph1}}
  \hfill
  \subfloat[Total Latency]{\includegraphics[width=0.4\columnwidth]{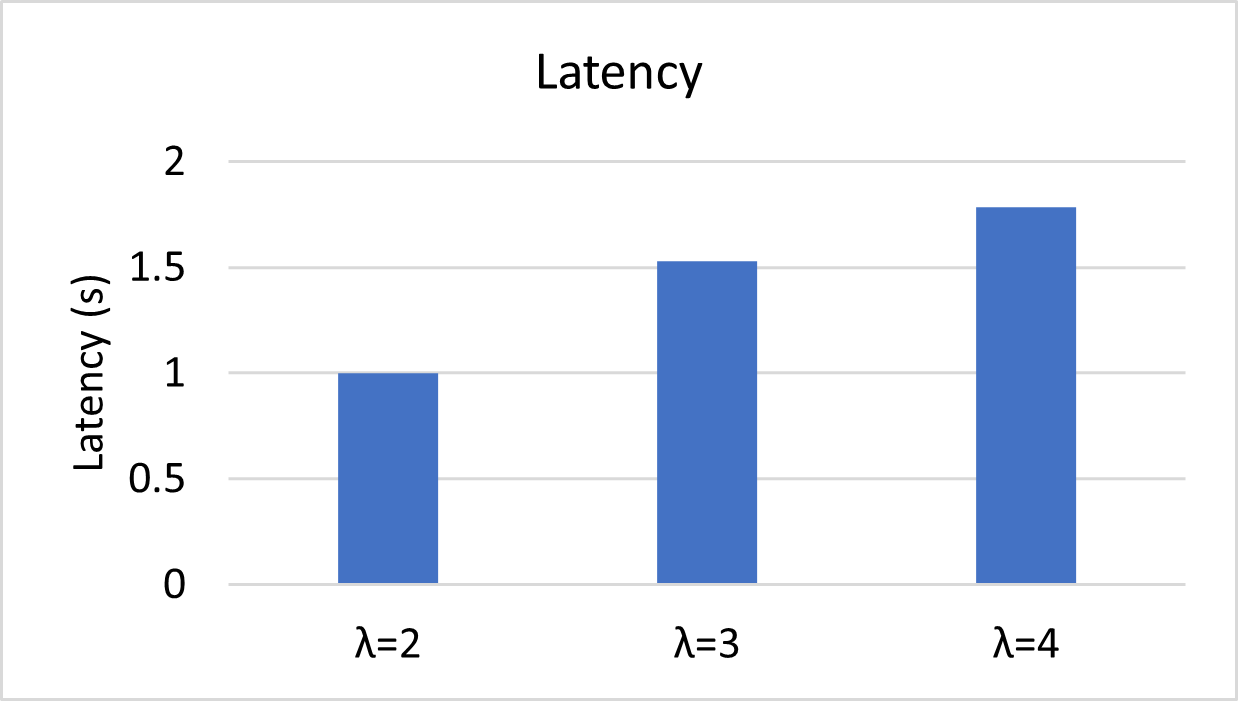}\label{figsc3:graph2}}
  \vfill
  \subfloat[Total Shared Data]{\includegraphics[width=0.4\columnwidth]{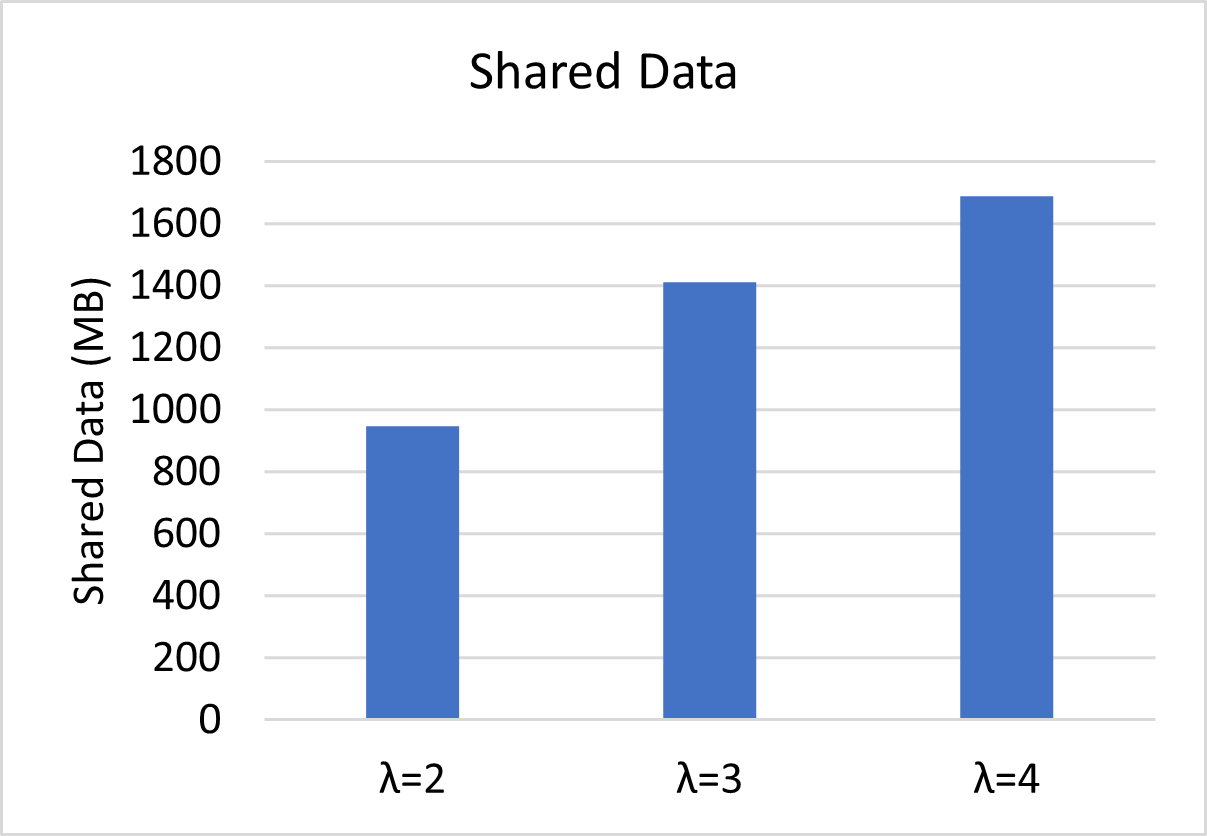}\label{figsc3:graph3}}
  \hfill
  \subfloat[Total Energy Consumption]{\includegraphics[width=0.4\columnwidth]{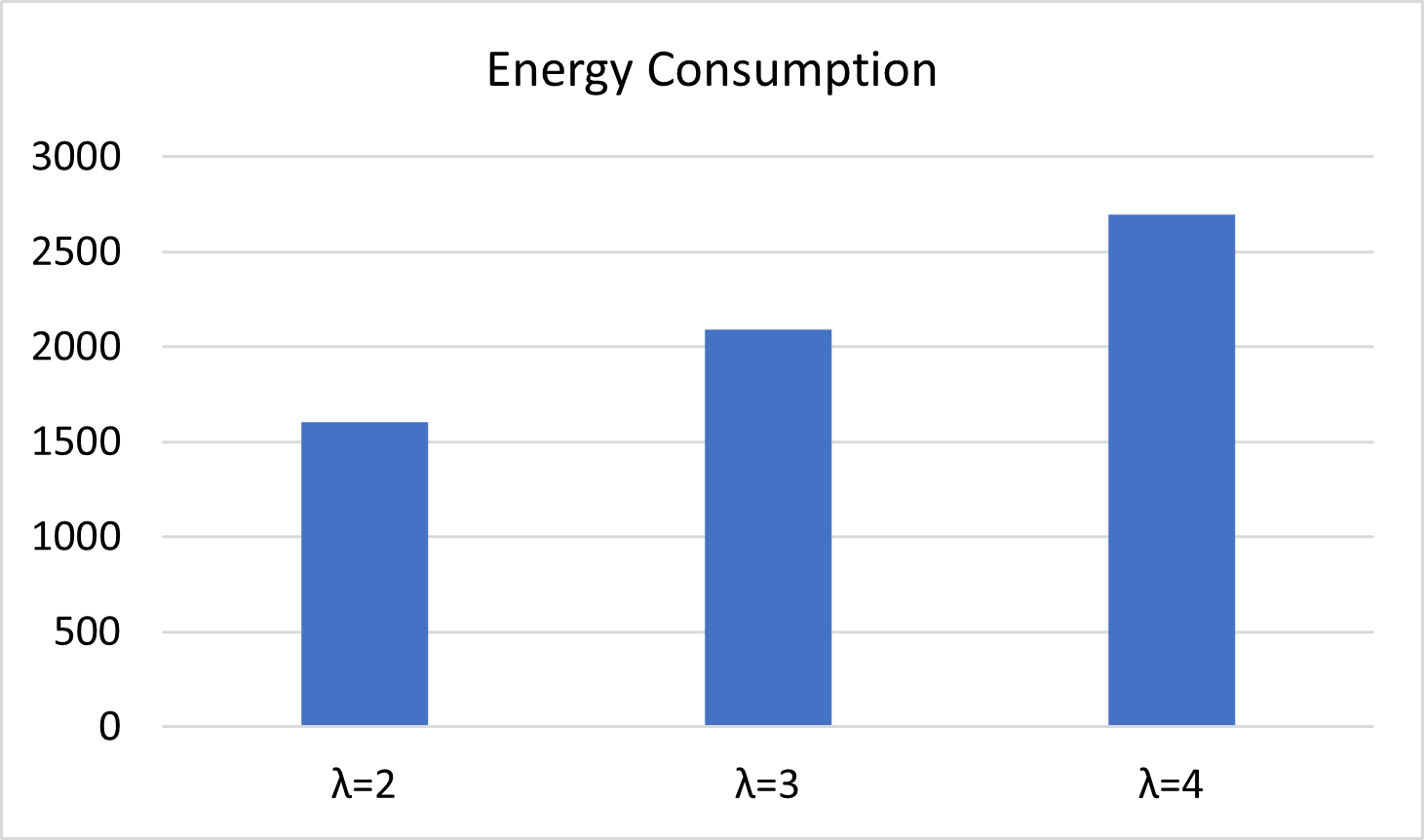}\label{figsc3:graph4}}
  \vfill
  \subfloat[Total Computation]{\includegraphics[width=0.4\columnwidth]{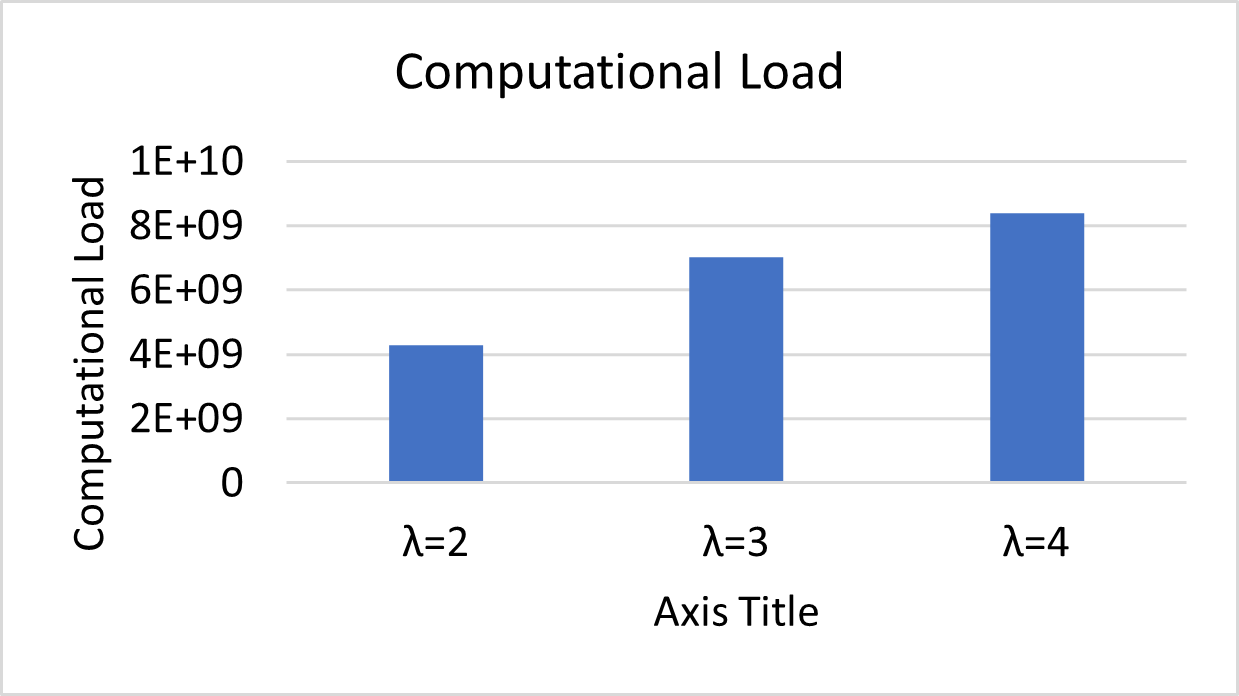}\label{figsc3:graph5}}
  \caption{Varying the rate of incoming requests}
  \label{fig:Sc3}
\end{figure}

While investigating this scenario we kept $\alpha$ = 0.7 and $\beta$ = 0.3. As increased $\lambda$ requires higher resources, the accuracy threshold is relaxed to 80\%. The energy of devices comprise of a combination of two types of devices with either 500J or 800J. This pushes the system to sacrifice the accuracy considering the resources are limited. Figure \ref{figsc3:graph1} shows the highest accuracy achieved where no blocks are dropped at the lowest data rate. It can be seen from figure \ref{figsc3:graph2} that the latency of the system is increasing for a similar accuracy due to the high rate of incoming requests. Additionally, with an increased $\lambda$,  figures \ref{figsc3:graph3}, \ref{figsc3:graph4} and \ref{figsc3:graph5} show an increased shared data, energy consumption and computational load.
 \\ 
\subsection{Investigating the change in Computational Capacity of the Devices:}
\begin{figure}[htbp]
  \centering
  \subfloat[Average Accuracy]{\includegraphics[width=0.45\columnwidth]{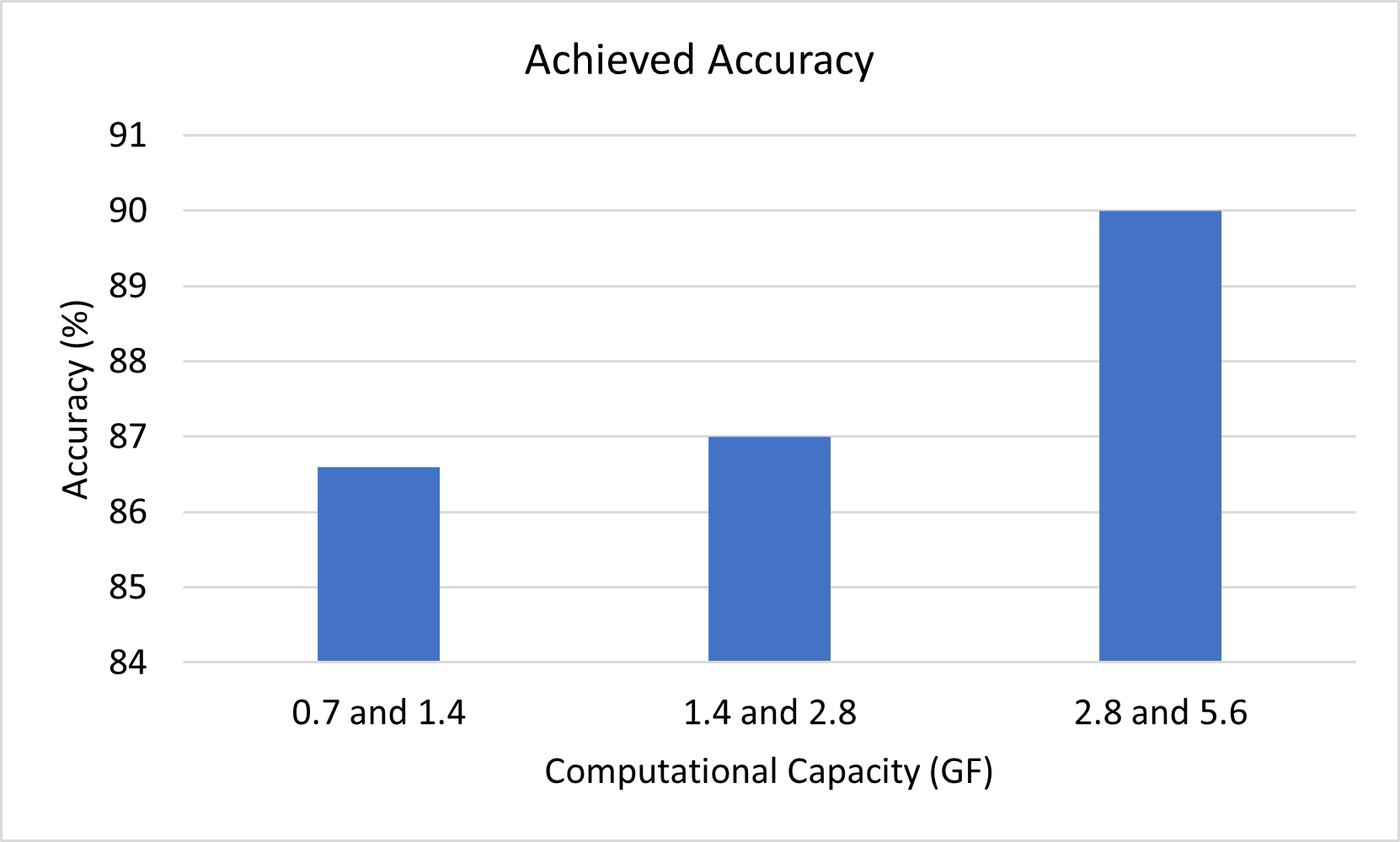}\label{fig:sc4graph1}}
  \hfill
  \subfloat[Total Latency]{\includegraphics[width=0.45\columnwidth]{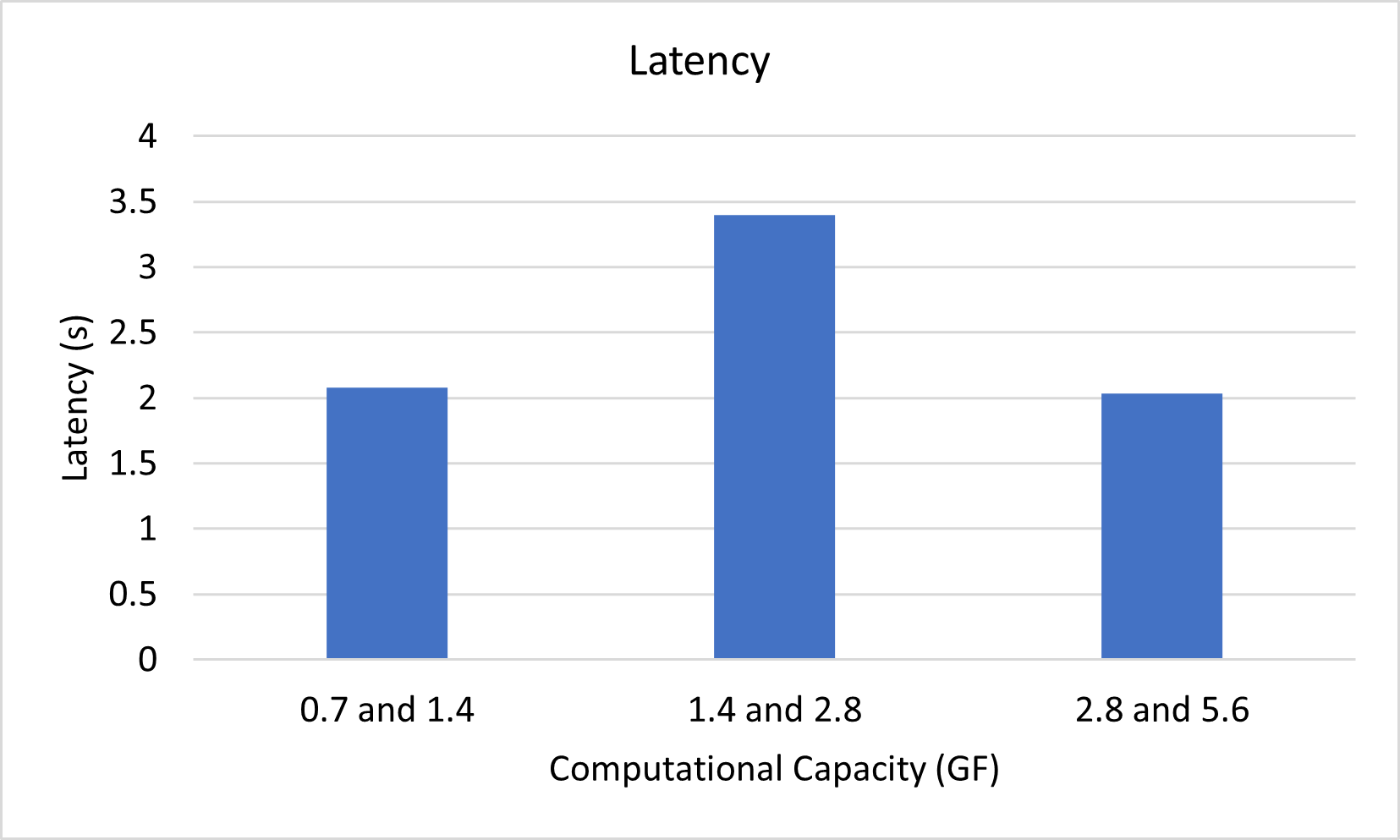}\label{figsc4:sc4graph2}}
  \vfill
  \subfloat[Total Shared Data]{\includegraphics[width=0.45\columnwidth]{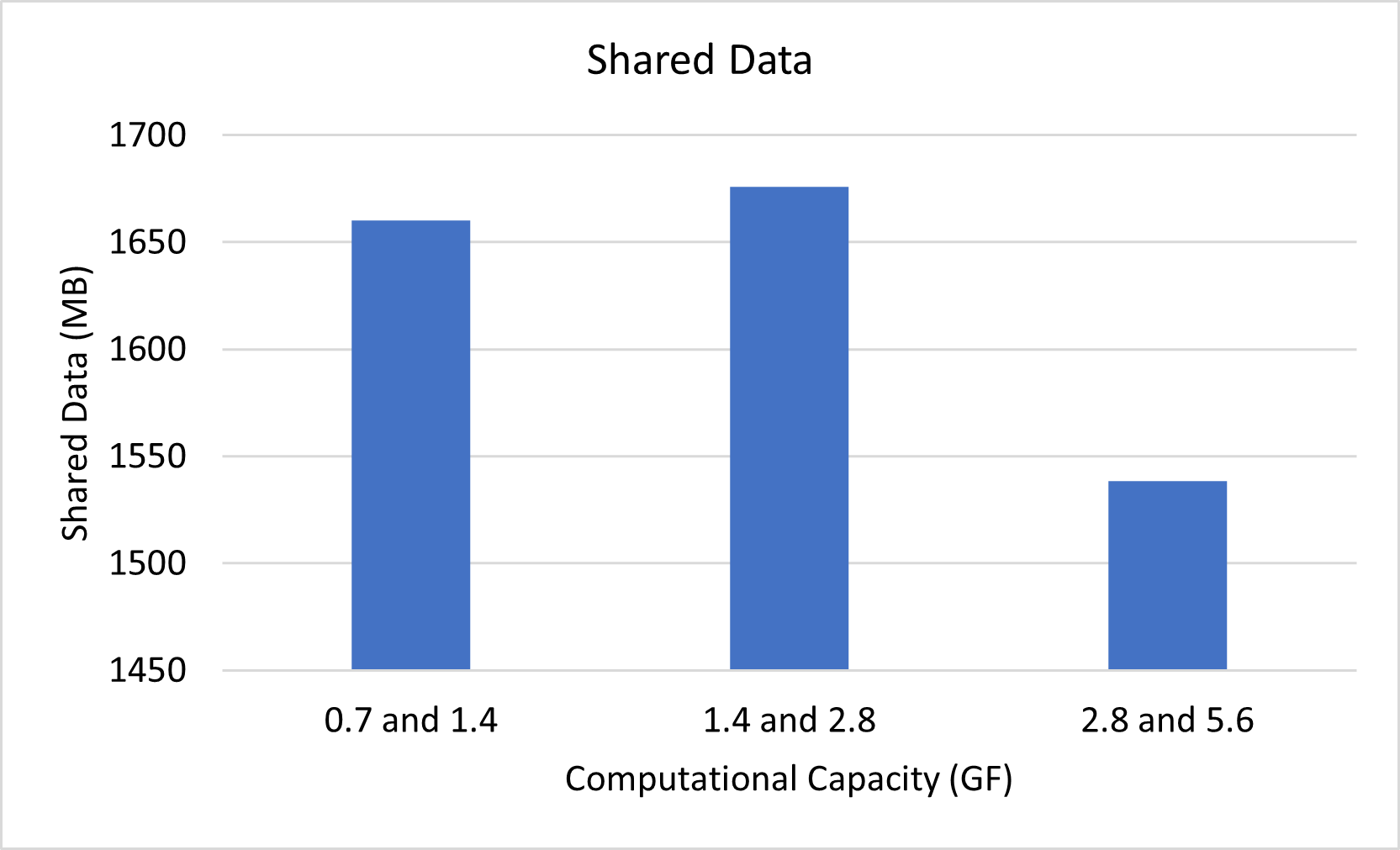}\label{figsc4:sc4graph3}}
  \hfill
  \subfloat[Total Energy Consumption]{\includegraphics[width=0.45\columnwidth]{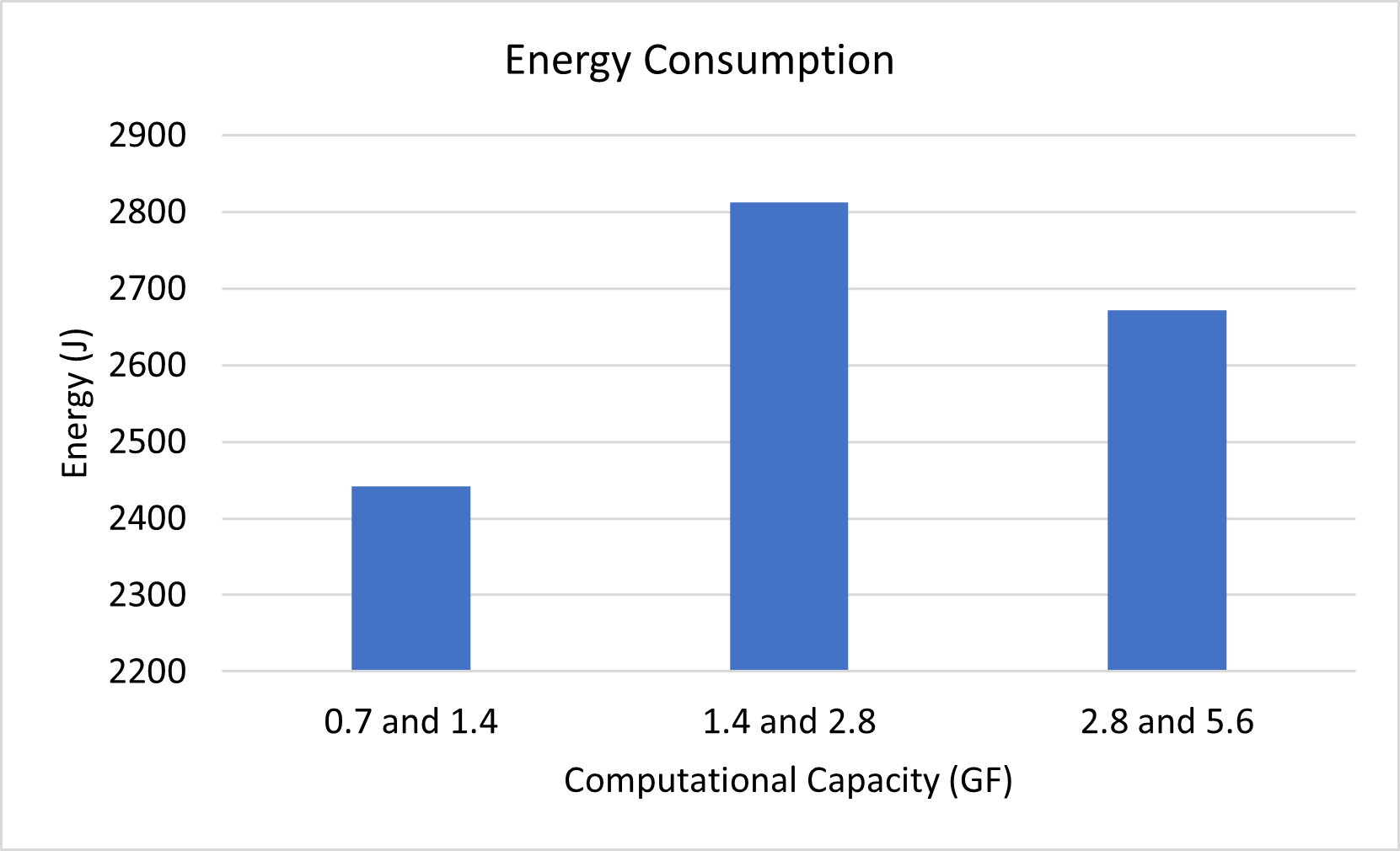}\label{figsc4:sc4graph4}}
  \caption{Varying Computational Capacity of the Devices}
  \label{fig:Sc4}
\end{figure}

This section investigates the performance when the computational capacity of the system is varied and, setting $\alpha$=0 and $\beta$=1 to prioritize the accuracy. We set the parameters as defined at the beginning of this section while keeping the energy of devices as 800J and 1000J. With a higher computational limit, the system is able to reach a higher accuracy as shown in figure \ref{fig:sc4graph1}. This illustrates the ability of the system to adapt the structure of the ResNet to the available resources and downsize the network when the computational resources are scarce. As per figure \ref{figsc4:sc4graph3}, the shared data increases as computational resources are low per device, hence each device delegates a part of the inference to other devices which increases the transmission latency and shared data. As per figures \ref{figsc4:sc4graph2} and \ref{figsc4:sc4graph4} we can identify the total latency and energy are lower for lowest computational capacity due higher downsizing.
\vspace{-5mm}

\section{CONCLUSION}
In conclusion, as a solution to utilize deep AI models in IoT devices, we propose to use distributed inference with ResNet architecture as it is more robust to failures. We propose an adaptive ResNet architecture that is motivated by the dynamics of the IoT network. We formulate a multi-objective optimization problem that targets minimizing the latency and maximizing the accuracy.  This is done by conducting an empirical study to identify the residual blocks in the ResNet-50 model that do not contribute significantly to the performance of the model. The results of this empirical study assist the optimization in reaching the lowest latency and highest accuracy as per the available resources while ensuring the network performance does deteriorate below a certain threshold. As a future work, we target to use the proposed system model with different datasets as well more ResNet structures. 
\section*{Acknowledgement}
This work was made possible by Qatar National Research Fund, NPRP grant NPRP13S-0205-200265. The findings achieved herein are solely the responsibility of the authors.
\bibliography{references.bib}
\bibliographystyle{unsrt}

\end{document}